\documentclass[11pt, a4paper, logo, copyright]{googlecloud}

\pdftrailerid{redacted}

\makeatletter
\renewcommand\bibentry[1]{\nocite{#1}{\frenchspacing\@nameuse{BR@r@#1\@extra@b@citeb}}}
\makeatother

\usepackage{kantlipsum, lipsum}
\usepackage{dsfont}

\usepackage[authoryear, sort&compress, round]{natbib}

\usepackage[utf8]{inputenc} 
\usepackage[T1]{fontenc}    
\usepackage{hyperref}       
\usepackage{url}            
\usepackage{booktabs}       
\usepackage{amsfonts}       
\usepackage{nicefrac}       
\usepackage{microtype}      
\usepackage{xcolor}         
\usepackage{tikzducks}
\usepackage{amsmath}
\usepackage{graphicx}
\usepackage{multirow}
\usepackage{makecell}
\usepackage{wrapfig}
\usepackage{colortbl}
\usepackage{pifont}
\usepackage{caption}
\usepackage{listings}
\usepackage[frozencache,cachedir=.]{minted}
\usepackage{algorithm}
\usepackage{algpseudocode}

\definecolor{myred}{rgb}{1, 0, 0}
\definecolor{myblue}{rgb}{0, 0, 1}
\definecolor{myblack}{rgb}{1, 1, 1}

\theoremstyle{plain}

\theoremstyle{definition}

\theoremstyle{remark}

\newcommand{\eg}{\emph{e.g.}}
\newcommand{\ie}{\emph{i.e.}}
\definecolor{Gray}{gray}{0.9}
\newcommand{\sname}{MLE-STAR}

\newcommand{\stdv}[2][\tiny]{#1\text{$\pm$}#2}

\title{MLE-STAR: Machine Learning Engineering Agent via Search and Targeted Refinement}

\correspondingauthor{jaehyun.nam@kaist.ac.kr, jinsungyoon@google.com}

\renewcommand{\today}{}

\author[1 2 *]{Jaehyun Nam}
\author[1]{Jinsung Yoon}
\author[1]{Jiefeng Chen}
\author[2]{Jinwoo Shin}
\author[1]{Sercan Ö. Arık}
\author[1]{Tomas Pfister}

\affil[1]{Google Cloud}
\affil[2]{KAIST}

\begin{abstract}
Agents based on large language models (LLMs) for machine learning engineering (MLE) can automatically implement ML models via code generation.
However, existing approaches to build such agents often rely heavily on inherent LLM knowledge and employ coarse exploration strategies that modify the entire code structure at once.
This limits their ability to select effective task-specific models and perform deep exploration within specific components, such as experimenting extensively with feature engineering options.
To overcome these, we propose \textit{\sname}, a novel approach to build MLE agents.
\sname~first leverages external knowledge by using a search engine to retrieve effective models from the web, forming an initial solution, then iteratively refines it by exploring various strategies targeting specific ML components.
This exploration is guided by ablation studies analyzing the impact of individual code blocks.
Furthermore, we introduce a novel ensembling method using an effective strategy suggested by \sname.
Our experimental results show that \sname~achieves medals in 64\% of the Kaggle competitions on the MLE-bench Lite, significantly outperforming the best alternative.
\end{abstract}

\begin{document}

\maketitle

\section{Introduction}
\label{sec:introduction}

\begin{figure*}[t]
\centering
\includegraphics[width=0.873\textwidth]{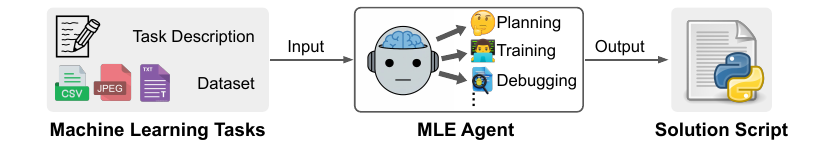}
\caption{
\textbf{Problem setup.} ML Engineering agents are designed to process a task description and datasets across various modalities (\eg, tabular, text, image, audio, etc.) with the objective of determining the optimal solution for a given machine learning problem, such as classification, regression, sequence-to-sequence generation, image denoising, text normalization, etc.
}
\label{fig:problem_setup}
\end{figure*}

The proliferation of machine learning (ML) has driven high-performance applications across diverse real-world scenarios, from fundamental tasks like tabular classification~\citep{chen2016xgboost, prokhorenkova2018catboost, hollmann2025accurate} to complex ones such as image denoising~\citep{fan2019brief}.
Despite these advances, developing such models remains a labor-intensive process for data scientists, involving extensive iterative experimentation and data engineering~\citep{hollmann2023large, nam2024optimized}.
To streamline such intensive workflows, recent research has focused on employing large language models (LLMs)~\citep{brown2020language, touvron2023llama, team2024gemini} as \textit{machine learning engineering (MLE) agents}~\citep{guo2024ds, hong2024data, jiang2025aide}.
By harnessing the coding and reasoning capabilities inherent in LLMs~\citep{jimenez2024swe, jain2025livecodebench}, these agents conceptualize ML tasks as code optimization problems. 
They then navigate the potential code solutions ultimately producing executable code (\eg, a Python script) based on a provided task description and dataset (see Figure~\ref{fig:problem_setup}).

Despite their promise as pioneering efforts, current MLE agents face several obstacles that limit their effectiveness.
First, due to their strong reliance on inherent LLM knowledge, they are often biased toward familiar and frequently used methods (\eg, the scikit-learn library~\citep{pedregosa2011scikit} for tabular data), neglecting potentially promising task-specific methods.
Additionally, these agents~\citep{jiang2025aide, guo2024ds} typically employ an exploration strategy that modifies the entire code structure at once in each iteration.
This often results in agents pivoting prematurely to other steps (\eg, model selection or hyperparameter tuning) because they lack the ability to perform deep, iterative exploration within specific pipeline components, such as experimenting different feature engineering options extensively.

\begin{figure*}[t]
\centering
\includegraphics[width=\textwidth]{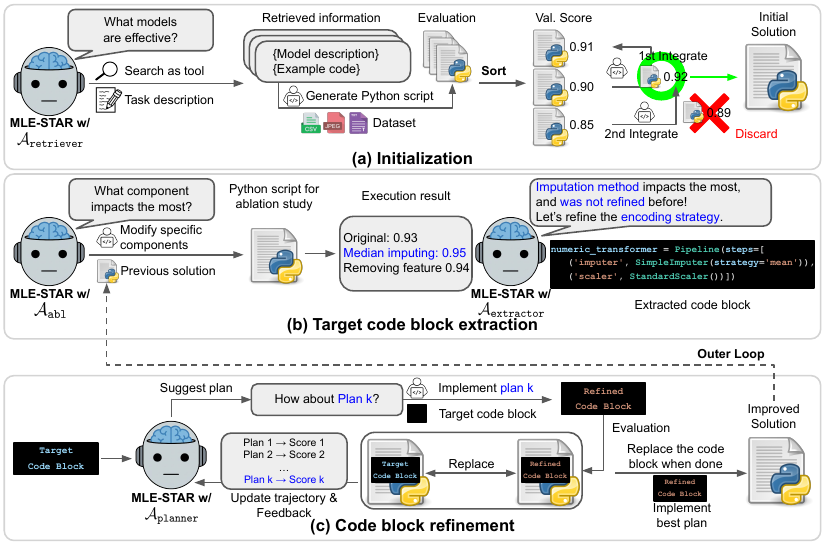}
\caption{
\textbf{Overview of \sname.} (a) Using search as a tool, \sname~retrieves task-specific models and uses them to generate an initial solution. (b) In each refinement step, \sname~performs an ablation study to extract the code block that have the greatest impact. Previously modified code blocks are also provided as feedback for diversity. (c) The extracted code block is iteratively refined based on plans suggested by the LLM, which explores various plans using previous experiments as feedback (\ie, inner loop), and the target code block is also selected repeatedly (\ie, outer loop, where the improved solution of (c) becomes the previous solution in (b)).
}
\label{fig:overview}
\end{figure*}

\noindent\textbf{Contributions.} We propose \textbf{\sname}, a novel \textbf{ML} \textbf{E}ngineering agent that integrates web \textbf{S}earch and \textbf{TA}rgeted code block \textbf{R}efinement (see Figure~\ref{fig:overview} for an overview).
Specifically, generating initial solution code, \sname~utilizes Google Search to retrieve relevant and potentially state-of-the-art approaches that could be effective towards building a model.
Moreover, to improve the solution, \sname~extracts a specific code block that represents a distinct ML pipeline component, such as feature engineering or ensemble building, and then concentrates on exploring strategies that are targeted to that component, using previous attempts as feedback to reflect on.
Here, to identify the code block that has the greatest impact on performance, \sname~performs an ablation study that evaluates the contribution of each ML component.
This refinement process is repeated, modifying various code blocks (\ie, other ML components).
In addition, we introduce a novel method to generate ensembles. \sname~first proposes multiple candidate solutions. Then, instead of relying on a simple voting based on validation scores, \sname~merges these candidates into a single improved solution using an ensemble strategy proposed by the agent itself. 
This ensemble strategy is iteratively refined based on the performance of the previous strategies.

To verify the effectiveness, we conduct comprehensive evaluations of \sname~using the MLE-bench's Kaggle competitions~\citep{chan2025mle}.
The experimental results demonstrate that \sname, requiring only minimal human effort (\eg, defining initial prompts that are generalizable to any tasks), significantly outperforms previous methods~\citep{jiang2025aide}, including those requiring manual labor to collect strategies from Kaggle~\citep{guo2024ds}.
In particular, \sname~achieves a substantial gain in medal achievement, improving it from 25.8\% to 63.6\% when compared to the top-performing baseline.
Additionally, we show that our proposed ensemble technique provides a meaningful improvement to \sname.
\section{Related work}\label{sec:related_work}

\textbf{LLM agents.}
Recent advances in LLMs have led to an active research in autonomous agents.
General-purpose agents like ReAct~\citep{yao2023react} and HuggingGPT~\citep{shen2023hugginggpt} typically use external tools to analyze various problems.
Specialized agents, such as Voyager~\citep{wang2023voyager} for Minecraft or AlphaCode~\citep{li2022competition} for code generation, excel in specific domains, often using execution feedback to iteratively improve their approach.
Extending these, we introduce \sname, an LLM agent that specialized in ML tasks.

\noindent\textbf{Automated machine learning.}
Automated machine learning (AutoML) aims to reduce reliance on human experts by automating end-to-end ML pipelines \citep{feurer2022auto, H2OAutoML20, jin2019auto}.
Auto-WEKA~\citep{kotthoff2017auto}, TPOT~\citep{olson2016tpot}, and recent advances such as AutoGluon~\citep{erickson2020autogluon}, have made progress through exploring within predefined model or hyperparameter spaces.
AutoML research also specializes in areas such as neural network design~\citep{pham2018efficient, zoph2016neural, real2019regularized, elsken2019neural}, and feature engineering~\citep{fan2010generalized, kanter2015deep, li2023learning, horn2019autofeat, zhang2023openfe}.
However, these methods rely on predefined search spaces, which often require domain expertise to define. To address this, LLM-based MLE agents~\citep{guo2024ds, jiang2025aide}, including \sname, are emerging, since they employ effective exploration strategies directly in the code space, without the need of manually-curated search spaces.

\noindent\textbf{MLE agents.}
Leveraging coding and reasoning capabilities of LLMs~\citep{jimenez2024swe, jain2025livecodebench}, research has been conducted on use of LLMs as MLE agents~\citep{hong2024data, schmidgall2025agent, li2024autokaggle}, which generate solution code, to automate ML workflows.
While MLAB~\citep{huang2024mlagentbench} and OpenHands~\citep{wang2024openhands} take general actions by calling tools to perform ML tasks, several studies specialize in ML automation.
AIDE~\citep{jiang2025aide} generates candidate solutions in a tree structure to facilitate code space exploration.
However, its heavy reliance on the LLM's internal knowledge can lead to outdated or overly simple model choices, and its refinement may prematurely shift focus between pipeline stages.
DS-Agent~\citep{guo2024ds} uses case-based reasoning~\citep{kolodner1992introduction, watson1994case} to discover strategies for solution generation by utilizing manually curated cases (primarily from Kaggle).
However, DS-Agent suffers from scalability issues due to its reliance on a manually built case bank, which requires significant human effort and can lead to solutions that are overfit to the source patterns.
Also, it restricts applicability to novel task types (like complex multi-modal problems).
Our method addresses these limitations.
Instead of attempting to explore the broader code space or relying on a static case bank, \sname~strategically explores implementation options for specific ML pipeline components.
It also improves scalability by using LLMs with search as tool to retrieve effective models that fit the task beyond the constraints of a fixed case bank.

\section{\sname}\label{sec:method}

We introduce the proposed framework for MLE agents, \sname, that effectively leverages the coding and reasoning capabilities of LLMs to solve ML tasks.
In a nutshell, our approach is based on first generating an initial solution by using web search as a tool (Section~\ref{subsec:initialization}), and then refining solutions via nested loops.
The outer loop targets one code block, which corresponds to the specific ML component extracted through an ablation study.
The inner loop iteratively refines \textit{only} this block until the outer loop moves to the next target (Section~\ref{subsec:refinement}).
We propose a novel ensemble method that improves the performance using the plan proposed by LLMs, which is iteratively refined (Section~\ref{subsec:ensemble}).
To mitigate potential undesirable behaviors from LLMs, such as using test sample statistics for missing value imputation, we introduce specific modules (detailed in Section~\ref{subsec:additional_modules}).
The prompts and algorithms used in each step can be found in Appendix~\ref{app:prompts} and \ref{app:algorithm}, respectively.

\noindent\textbf{Problem setup.}
Formally, our goal is to find an optimal solution $s^{*}=\arg\max_{s\in\mathcal{S}}h(s)$, where $\mathcal{S}$ is the space of possible solutions (\ie, Python scripts) and $h: \mathcal{S}\rightarrow\mathbb{R}$ is a score function (\eg, validation accuracy)~\citep{jiang2025aide}.
To obtain $s^{*}$, we propose a multi-agent framework $\mathcal{A}$, which takes datasets $\mathcal{D}$ (that might contain multiple files) and a task description $\mathcal{T}_{\mathtt{task}}$ (which includes task types, data modalities, score functions, etc.) as input.\footnote{\sname~works across any data modalities (\eg, tabular, image, text, audio) and task types (\eg, classification, image-to-image, sequence-to-sequence) -- it is not restricted to specific inputs or objectives.} 
Here, $\mathcal{A}$ consists of $n$ LLM agents $(\mathcal{A}_1, \cdots, \mathcal{A}_n)$. Each agent $\mathcal{A}_i$ possesses specific functionalities, which are elaborated upon in following sections.

\subsection{Generating an initial solution using web search as a tool}\label{subsec:initialization}

\textbf{Candidate model search.} \sname~starts by generating an initial solution.
For high performance in ML tasks, selecting the appropriate model is paramount.
However, relying solely on an LLM for model suggestions can lead to suboptimal choices.
For instance, we observe that LLMs propose models like logistic regression~\citep{pedregosa2011scikit} even for competitions like jigsaw-toxic-comment-classification, which is a text classification task, potentially because LLMs favor familiar patterns from their pre-training data over up-to-date information.
To mitigate this, we propose using web search as a tool for \sname~first to retrieve $M$ effective, state-of-the-art models for the given task.
This retrieved context is then used to guide the LLM in generating a more informed initial solution. Formally:
\begin{equation}
    \{\mathcal{T}^{i}_{\mathtt{model}}, \mathcal{T}^{i}_\mathtt{code}\}_{i=1}^{M} = \mathcal{A}_{\mathtt{retriever}} (\mathcal{T}_\mathtt{task}),
\end{equation}
where $\mathcal{T}_\mathtt{model}$ represents the description of a retrieved model, while $\mathcal{T}_\mathtt{code}$ provides corresponding example code.
This example code is needed since the LLM can be unfamiliar with the model and cannot generate the executable code without proper guidance.
Then, \sname~involves evaluating of the performance of model $i$.
To achieve this, candidate evaluation agent $\mathcal{A}_\mathtt{init}$ first generates code, $s_\mathtt{init}^i$, using the retrieved model to solve the given ML task.
This process is formally defined as:
\begin{equation}
    s_\mathtt{init}^i = \mathcal{A}_\mathtt{init} (\mathcal{T}_\mathtt{task}, \mathcal{T}_\mathtt{model}^i, \mathcal{T}_\mathtt{code}^i).
\end{equation}
We evaluate the performance of each $s$ using a task-specific metric $h$ on dataset $\mathcal{D}$. We denote the resulting score by $h(s)$, which encapsulates the entire process done in $s$: splitting $\mathcal{D}$ into training and validation sets, training the model specified in $s$ using the training data, and calculating $h$ on the validation data. The performance for $s_\mathtt{init}^i$ is thus $h(s_\mathtt{init}^i)$. 
As a result, a set of code scripts $\mathcal{S}_\mathtt{init}=\{s_\mathtt{init}^1, \cdots, s_\mathtt{init}^M\}$ and their performance scores $\{h(s_\mathtt{init}^1), \cdots, h(s_\mathtt{init}^M)\}$ are obtained.

\noindent\textbf{Merging candidate models for initial solution.}
After the evaluation of the $M$ retrieved models, a consolidated initial solution $s_0$ is constructed through an iterative merging procedure. Specifically, we first define $\pi$ be a permutation of the indices such that the scores are sorted in descending order: $h(s_\mathtt{init}^{\pi(1)})\geq h(s_\mathtt{init}^{\pi(2)}) \geq \cdots \geq h(s_\mathtt{init}^{\pi(M)})$.
Then, we initialize the initial solution $s_0$ with the top-performing script, and record the current best score, \ie, $s_0\leftarrow s_{(1)}$, $h_\mathtt{best}\leftarrow h(s_0)$,
where $s_{(k)}$ denote the script $s_\mathtt{init}^{\pi(k)}$ for simplicity.
Finally, we sequentially attempt to incorporate the remaining scripts $s_{(k)}$ for $k = 2, \cdots, M$ into $s_0$. For each $k$, \sname~creates a candidate merged script by leveraging an agent $\mathcal{A}_\mathtt{merger}$ that attempts to integrate $s_{(k)}$ into the current $s_0$. Formally,
\begin{equation}
    s_0\leftarrow\mathcal{A}_\mathtt{merger}(s_0, s_{(k)}),~h_\mathtt{best}\leftarrow h(s_0)
\end{equation}
where, $\mathcal{A}_\mathtt{merger}$ is guided to introduce a simple average ensemble to merge multiple models.
Finally, we merge the models until the validation score $h_\mathtt{best}$ no longer improves (see Appendix~\ref{app:algorithm}).



\subsection{Refining a code block for solution improvement}\label{subsec:refinement}

The iterative refinement phase begins with an initial solution $s_0$ and proceeds for a predetermined number of $T$ outer loop steps, indexed by $t=0,1,\cdots,T-1$. At each step $t$, the goal is to improve the current solution $s_t$ to obtain $s_{t+1}$, optimizing for a performance metric $h$. This process involves two main stages: targeted code block extraction and code block refinement.

\noindent\textbf{Targeted code block extraction.}
To effectively explore specialized improvement strategies, \sname~identifies and targets specific code blocks within the ML pipeline represented by $s_t$. This selection is guided by an ablation study performed by an agent $\mathcal{A}_\mathtt{abl}$.
Specifically, the agent $\mathcal{A}_\mathtt{abl}$ generates a code $a_t$ designed to perform an ablation study on $s_t$. This script creates variations of $s_t$ by modifying or disabling specific components. To encourage exploration of different pipeline parts across iterations, $\mathcal{A}_\mathtt{abl}$ receives the summaries of previous ablation studies $\{\mathcal{T}_\mathtt{abl}^i\}_{i=0}^{t-1}$ as input:
\begin{equation}
    a_t=\mathcal{A}_\mathtt{abl}(s_t, \{\mathcal{T}_\mathtt{abl}^i\}_{i=0}^{t-1}).
\end{equation}
Then, $a_t$ is executed, producing output results $r_t$ (see examples in Appendix~\ref{app:qual_ex}). A summarization module $\mathcal{A}_\mathtt{summarize}$ processes the script and its results to generate a concise ablation summary $\mathcal{T}_\mathtt{abl}^t$:
\begin{equation}
    \mathcal{T}_\mathtt{abl}^t = \mathcal{A}_\mathtt{summarize}(a_t, r_t).
\end{equation}
Finally, \sname~selects a critical block which has the most significant impact on performance. To achieve this, an extractor module $\mathcal{A}_\mathtt{extractor}$ analyzes the $\mathcal{T}_\mathtt{abl}^t$ to identify the code block $c_t$ within $s_t$ whose modification had the most significant impact on performance. To prioritize refinement of blocks not previously targeted, the set of already refined blocks $\{c_i\}_{i=0}^{t-1}$ is provided as context:
\begin{equation}\label{eq:extract}
    c_t, p_0=\mathcal{A}_\mathtt{extractor}(\mathcal{T}_\mathtt{abl}^t, s_t, \{c_i\}_{i=0}^{t-1}).
\end{equation}
Here, \sname~also generates the initial plan $p_0$ for code block refinement at the same time, since $\mathcal{T}_\mathtt{abl}$ can provide a good starting point by analyzing the modification of corresponding component.

\noindent\textbf{Code block refinement.}
Once the targeted code block $c_t$ is defined, \sname~explores various refinement strategies to improve the metric $h$.
This involves an inner loop exploring $K$ potential refinement for $c_t$.
An agent $\mathcal{A}_\mathtt{coder}$ first implements $p_0$, transforming $c_t$ into a refined block $c_t^0$, \ie, $c_t^0=\mathcal{A}_\mathtt{coder}(c_t,p_0)$.
A candidate solution $s_t^0$ is formed by substituting $c_t^0$ into $s_t$:
\begin{equation}
    s_t^0=s^t.\mathtt{replace}(c_t,c_t^0),
\end{equation}
where, $\mathtt{replace}$ denotes the code replacement operation.
Finally, the performance $h(s_t^0)$ is evaluated.

To discover potentially more effective or novel refinement strategies, \sname~iteratively generates and evaluates further plans. For $k=1,\cdots,K-1$, a planning agent $\mathcal{A}_\mathtt{planner}$ proposes the next plan $p_k$.
This agent leverages the previous attempts within the current outer step $t$ as feedback:
\begin{equation}
    p_k=\mathcal{A}_\mathtt{planner}(c_t, \{(p_j, h(s_t^j))\}_{j=0}^{k-1}).
\end{equation}
For each plan $p_k$, the coding agent generates the corresponding refined block, \ie, $c_t^k=\mathcal{A}_\mathtt{coder}(c_t,p_k)$, creates the candidate solution $s_t^k=s_t.\mathtt{replace}(c_t, c_t^k)$, and evaluates its performance $h(s_t^k)$.
After exploring $K$ refinement strategies (indexed $k=0,\cdots,K-1$), the best-performing candidate solution is identified: $k^*=\arg\max_{k\in\{0,\cdots,K-1\}}h(s_t^k)$. The solution for the next outer step, $s_{t+1}$, is updated to $s_{t}^{k^*}$ only if an improvement over $s_t$ is found.
This iterative process continues until $t=T$.

\begin{figure*}[t]
\centering
\includegraphics[width=\textwidth]{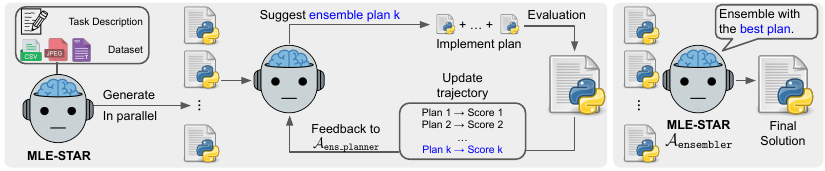}
\caption{
\textbf{Ensembling solutions.} \sname~iteratively proposes effective ensemble strategies based on previous attempts, integrating multiple solutions generated in parallel into a single solution.
}
\label{fig:tts}
\end{figure*}
\subsection{Further improvement by exploring ensemble strategies}\label{subsec:ensemble}
To further improve upon the best single solution generated, we introduce a novel ensembling procedure (Figure~\ref{fig:tts}).
Standard practice might involve generating multiple candidate solutions and selecting the one with the highest score~\citep{ichihara2025evaluation} according to metric $h$.
However, analogous to model ensembling, we posit that suboptimal solutions might contain complementary strengths, and combining multiple solutions could lead to superior performance compared to relying on any single one.
Therefore, we employ the planning capabilities of \sname~to automatically discover effective strategies for ensembling.
Specifically, let $\{s_l\}_{l=1}^L$ be a set of $L$ distinct solutions obtained (\eg, from parallel runs of the process described earlier).
Our goal is to find an effective ensemble plan $e$ that merges these solutions, which mirrors the structure of the targeted code block refinement stage.
We start with an initial ensemble plan $e_0$ (\eg, a simple strategy like averaging the final predictions obtained from the models trained using each solution $s_l$), proposed by \sname~itself.
After the performance $h(s_\mathtt{ens}^0)$ for the initial plan $e_0$ is calculated, for a fixed number of iterations, $r=1,\cdots,R$, the planning agent $\mathcal{A}_\mathtt{ens\_planner}$, specialized in suggesting ensemble plans, proposes subsequent ensemble plans $e_r$.
This agent uses the history of previously attempted ensemble plans and their resulting performance as feedback, \ie, $e_r = \mathcal{A}_\mathtt{ens\_planner}(\{s_l\}_{l=1}^{L},\{(e_j, h(s_\mathtt{ens}^j)) \}_{j=0}^{r-1})$.
Each $e_r$ is implemented via $\mathcal{A}_\mathtt{ensembler}$ to obtain $s_\mathtt{ens}^r$:
\begin{equation}
    s_\mathtt{ens}^r = \mathcal{A}_\mathtt{ensembler}(e_r, \{s_l\}_{l=1}^{L}). 
\end{equation}
Finally, after exploring $R$ ensemble strategies, the ensemble result that achieves the highest performance is selected as the final output, yielding the final ensembled result $s_\mathtt{ens}^*=s_\mathtt{ens}^{r^*}$: $r^* = \arg\max_{r \in \{0, \dots, R\}} h(s_\mathtt{ens}^r)$.
This procedure allows \sname~to autonomously explore and identify potentially novel and effective ways to combine multiple complex solutions.

\subsection{Additional modules for robust MLE agents}\label{subsec:additional_modules}

\textbf{Debugging agent.}
We detail the design of our debugging agent within \sname.
If the execution of a Python script $s$ triggers an error, resulting in a record $\mathcal{T}_\mathtt{bug}$ (\eg, a traceback), \sname~employs a debugging module $\mathcal{A}_\mathtt{debugger}$ to attempt correction.
This process iteratively updates the script:
\begin{equation}
    s\leftarrow\mathcal{A}_\mathtt{debugger}(s,\mathcal{T}_\mathtt{bug}).
\end{equation}
The debugging step is repeated until either the script executes successfully, or a predefined maximum number of debugging rounds is reached.
If the bug cannot be resolved, \sname~ proceeds to the next task using the latest version of the script that is known to be executable.

\noindent\textbf{Data leakage checker.}
We observe that LLM-generated Python scripts might have the risk of introducing data leakage, for example, by improperly accessing information from a test dataset during training dataset preparation (see Figure~\ref{fig:data_leakage_checker}).
To address this, we introduce a checker agent, $\mathcal{A}_\mathtt{leakage}$, which analyzes the solution script $s$ prior to its execution.
Recognizing that full-script analysis can be inefficient for lengthy code, we adopt a targeted approach.
First, we extract the code block $c_\mathtt{data}$ where data preprocessing is done.
Second, $c_\mathtt{data}$ is passed to the checker.
If $\mathcal{A}_\mathtt{leakage}$ detects potential data leakage, it generates a corrected version $c_\mathtt{data}^{*}$: $c_\mathtt{data}^{*}=\mathcal{A}_\mathtt{leakage}(c_\mathtt{data})$.
Finally, the original script $s$ is updated by replacing the identified segment with its corrected version: 
$s\leftarrow s.\mathtt{replace}(c_\mathtt{data}, c_\mathtt{data}^{*})$.
If no leakage is detected in $c_\mathtt{data}$ by $\mathcal{A}_\mathtt{leakage}$, the script $s$ remains unmodified. All generated solutions are passed through a data leakage checker, $\mathcal{A}_\mathtt{leakage}$, prior to their execution for evaluation.

\noindent\textbf{Data usage checker.}
We observe that LLM-generated scripts sometimes neglect using provided data sources, focusing solely on simple formats like CSVs (see Figure~\ref{fig:data_usage_checker}).
To ensure the utilization of all relevant provided data, \sname~introduces a data usage checker agent, $\mathcal{A}_\mathtt{data}$. Specifically, before \sname~starts refinement, $\mathcal{A}_\mathtt{data}$ checks the initial solution $s_0$ along with the task description $\mathcal{T}_\mathtt{task}$.
If relevant provided data is not adequately used, $\mathcal{A}_\mathtt{data}$ revises the initial script as:
\begin{equation}
    s_0\leftarrow\mathcal{A}_\mathtt{data}(s_0, \mathcal{T}_\mathtt{task}).
\end{equation}
\section{Experiments}\label{sec:experiments}

In this section, we validate the effectiveness of \sname~using 22 Kaggle competitions from MLE-bench Lite~\citep{chan2025mle}. Our results demonstrate that \sname~significantly outperforms baselines, including those employing various LLMs (Section~\ref{subsec:main_results}).
Furthermore, we show that using better models and leveraging our proposed ensemble strategy effectively improves performance (Section~\ref{subsec:ablations}).
We also provide the example solutions generated by \sname, in Appendix~\ref{app:full_code}.

\noindent\textbf{Common setup.}
All experiments are conducted on 22 Kaggle competitions from MLE-bench Lite~\citep{chan2025mle} using three random seeds and Gemini-2.0-Flash, unless otherwise specified.
Here, we use an agent $\mathcal{A}_\mathtt{test}$, which takes the task description and the final solution as input, and outputs the code that incorporates loading test sample and creating a submission file (see Appendix~\ref{app:benchmark} for details).
\sname~begins by retrieving four model candidates.
\sname~refines for four inner loops, while exploring four outer loops.
For ensemble, \sname~generates two solutions in parallel, and explore ensemble strategies for five rounds.
Following the MLE-bench's setup, we set a maximum time limit of 24 hours for a fair comparison (see computation analysis in Appendix~\ref{app:exp}).
We primarily consider AIDE~\citep{jiang2025aide} as our main baseline, given its state-of-the-art performance on MLE-bench.
It is important to note that other baselines often limit their generalizability across various task types (\eg, audio classification, sequence-to-sequence), frequently showcasing results only on simpler modalities like tabular~\citep{hong2024data, li2024autokaggle}.
For instance, DS-Agent~\citep{guo2024ds} requires a manually constructed case bank, and their current GitHub repository lacks cases for audio classification, sequence-to-sequence, image classification, etc.

\begin{table*}[t]
\caption{\textbf{Main results from MLE-bench Lite.} Each experiment is repeated using three seeds, except for o1-preview (AIDE) and GPT-4o (AIDE), which use 16 and 36 seeds, respectively. All results are taken from the GitHub repository of MLE-bench paper~\citep{chan2025mle}, except for the model using Gemini. Scores represent the mean and one standard error of the mean.}\label{tab:main}
\vspace{-0.3in}
\begin{center}
\resizebox{1.0\textwidth}{!}{
\begin{tabular}{lccccccc}
\toprule
Model & \begin{tabular}[c]{c}Made\\Submission\\ (\%) \end{tabular} & \begin{tabular}[c]{c}Valid\\Submission\\ (\%) \end{tabular} & \begin{tabular}[c]{c}Above\\Median\\ (\%) \end{tabular} & \begin{tabular}[c]{c}Bronze\\ (\%) \end{tabular} & \begin{tabular}[c]{c}Silver\\ (\%) \end{tabular} & \begin{tabular}[c]{c}Gold\\ (\%) \end{tabular} & \cellcolor{Gray} \begin{tabular}[c]{c}Any\\Medal\\ (\%) \end{tabular} \\

\midrule

\multicolumn{8}{l}{\textbf{\sname~(Ours)}}\\
\midrule
\textbf{gemini-2.5-pro} & \textbf{100.0}\stdv{0.0} & \textbf{100.0}\stdv{0.0} & \textbf{83.3}\stdv{4.6}& 6.1\stdv{3.0} & \textbf{21.2}\stdv{5.1} & \textbf{36.4}\stdv{6.0} & \cellcolor{Gray}\textbf{63.6}\stdv{6.0}\\
gemini-2.0-flash & \phantom{0}95.5\stdv{2.6} & \phantom{0}95.5\stdv{2.6} & 63.6\stdv{6.0} & \textbf{9.1}\stdv{3.6} & \phantom{0}4.5\stdv{2.6} & 30.3\stdv{5.7} & \cellcolor{Gray}43.9\stdv{6.2} \\
\midrule
\multicolumn{8}{l}{\textbf{AIDE~\citep{jiang2025aide}}}\\
\midrule
gemini-2.0-flash & \phantom{0}87.9\stdv{4.0} & \phantom{0}78.8\stdv{5.0} & 39.4\stdv{6.0} & 4.5\stdv{2.6} & \phantom{0}9.1\stdv{3.5} & 12.1\stdv{4.0} & \cellcolor{Gray}25.8\stdv{5.4}\\
o1-preview & \phantom{0}99.7\stdv{0.3} & \phantom{0}90.3\stdv{1.6} & 58.2\stdv{2.6} & 4.8\stdv{1.1} & 11.1\stdv{1.7} & 20.7\stdv{2.2} & \cellcolor{Gray}36.6\stdv{2.6}\\
gpt-4o & \phantom{0}82.1\stdv{1.4} & \phantom{0}65.7\stdv{1.7} & 29.9\stdv{1.6} & 3.4\stdv{0.6} & \phantom{0}5.8\stdv{0.8} & \phantom{0}9.3\stdv{1.0} & \cellcolor{Gray}18.6\stdv{1.4}\\
llama-3.1-405b-instruct & \phantom{0}72.7\stdv{5.5} & \phantom{0}51.5\stdv{6.2} & 18.2\stdv{4.7} & 0.0\stdv{0.0} & \phantom{0}4.5\stdv{2.6} & \phantom{0}6.1\stdv{2.9} & \cellcolor{Gray}10.6\stdv{3.8}\\
claude-3-5-sonnet & \phantom{0}81.8\stdv{4.7} & \phantom{0}66.7\stdv{5.8} & 33.3\stdv{5.8} & 3.0\stdv{2.1} & \phantom{0}6.1\stdv{2.9} & 10.6\stdv{3.8} & \cellcolor{Gray}19.7\stdv{4.9}\\
\midrule
\multicolumn{8}{l}{\textbf{MLAB~\citep{huang2024mlagentbench}}}\\
\midrule
gpt-4o & \phantom{0}84.8\stdv{4.4} & \phantom{0}63.6\stdv{5.9} & \phantom{0}7.6\stdv{3.3} & 3.0\stdv{2.1} & \phantom{0}1.5\stdv{1.5} & \phantom{0}1.5\stdv{1.5} & \cellcolor{Gray}\phantom{0}6.1\stdv{2.9}\\
\midrule
\multicolumn{8}{l}{\textbf{OpenHands~\citep{wang2024openhands}}}\\
\midrule
gpt-4o & \phantom{0}81.8\stdv{4.7} & \phantom{0}71.2\stdv{5.6} & 16.7\stdv{4.6} & 3.0\stdv{2.1} & \phantom{0}3.0\stdv{2.1} & \phantom{0}6.1\stdv{2.9} & \cellcolor{Gray}12.1\stdv{4.0}\\
\bottomrule
\end{tabular}
}
\end{center}
\end{table*}
\begin{table*}[t]
\begin{minipage}[b]{0.46\textwidth} 
\centering
\caption{Comparison with DS-Agent.}\label{tab:ds-agent}
\vspace{-0.1in}
\small
\begin{tabular}{cccc}
\toprule
Task & Metric & DS-Agent & \textbf{\sname}\\
\midrule
WBY & MAE ($\downarrow$) & 213 & \textbf{166}\\
MCC & RMLSE ($\downarrow$) & 0.2964 & \textbf{0.2911}\\
ST & Accuracy ($\uparrow$) & 0.7982 & \textbf{0.8091}\\
ES & AUROC ($\uparrow$) & 0.8727 & \textbf{0.9101}\\
\bottomrule
\end{tabular}
\end{minipage} \hfill 
\begin{minipage}[b]{0.52\textwidth} 
\centering
\small
\caption{Performance with Claude-Sonnet-4.}\label{tab:gemini-2.5}
\vspace{-0.1in}
\begin{tabular}{cccc}
\toprule
Task & Metric & Gemini-2.0-flash & \textbf{Sonnet 4}\\
\midrule
DDD & RMSE ($\downarrow$) & 0.0681 & \textbf{0.0155}\\
DBI & Log Loss ($\downarrow$) & 0.4535 & \textbf{0.3114}\\
SAI & Log Loss ($\downarrow$) & 0.2797 & \textbf{0.2610} \\
WCR & AUROC ($\uparrow$) & \textbf{0.9903} & 0.9888\\
\bottomrule
\end{tabular}
\end{minipage}
\end{table*}
\begin{table*}[t]
\caption{\textbf{Ablation on ensemble strategy.} Experiment results on MLE-bench Lite, repeated three seeds using Gemini-2.0-Flash. Scores represent the mean and one standard error of the mean.}\label{tab:abl_tts}
\vspace{-0.3in}
\begin{center}
\resizebox{1.0\textwidth}{!}{
\begin{tabular}{lcccccccc}
\toprule
Ensemble strategy & \begin{tabular}[c]{c}Made\\Submission\\ (\%) \end{tabular} & \begin{tabular}[c]{c}Valid\\Submission\\ (\%) \end{tabular} & \begin{tabular}[c]{c}Above\\Median\\ (\%) \end{tabular} & \begin{tabular}[c]{c}Bronze\\ (\%) \end{tabular} & \begin{tabular}[c]{c}Silver\\ (\%) \end{tabular} & \begin{tabular}[c]{c}Gold\\ (\%) \end{tabular} & \begin{tabular}[c]{c}Any\\Medal\\ (\%) \end{tabular} \\

\midrule

\multicolumn{8}{l}{\textbf{AIDE~\citep{jiang2025aide}}}\\

\midrule
None & 87.9\stdv{4.0} & 78.8\stdv{5.0} & 39.4\stdv{6.0} & 4.5\stdv{2.6} & \phantom{0}9.1\stdv{3.5} & 12.1\stdv{4.0} & 25.8\stdv{5.4}\\

\midrule

\multicolumn{8}{l}{\textbf{\sname~(Ours)}}\\

\midrule

None & \textbf{95.5}\stdv{2.6} & \textbf{95.5}\stdv{2.6} & 57.6\stdv{6.1} & 7.6\stdv{3.3} & \phantom{0}4.5\stdv{2.6} & 25.8\stdv{5.4} & 37.9\stdv{6.0} \\
Best-of-N & \textbf{95.5}\stdv{2.6} & \textbf{95.5}\stdv{2.6} & 62.1\stdv{6.0} & 6.1\stdv{3.0} & \phantom{0}7.6\stdv{3.3} & 28.8\stdv{5.6} & 42.4\stdv{6.1} \\
Average ensemble & \textbf{95.5}\stdv{2.6} & \textbf{95.5}\stdv{2.6} & 60.6\stdv{6.1} & 6.1\stdv{3.0} & \textbf{12.1}\stdv{4.0} & 25.8\stdv{9.4} & \textbf{43.9}\stdv{6.2} \\
\cellcolor{Gray}\textbf{Ours} & \cellcolor{Gray}\textbf{95.5}\stdv{2.6} & \cellcolor{Gray}\textbf{95.5}\stdv{2.6} & \cellcolor{Gray}\textbf{63.6}\stdv{6.0} & \cellcolor{Gray}\textbf{9.1}\stdv{3.6} & \cellcolor{Gray}\phantom{0}4.5\stdv{2.6} & \cellcolor{Gray}\textbf{30.3}\stdv{5.7} & \cellcolor{Gray}\textbf{43.9}\stdv{6.2} \\

\bottomrule
\end{tabular}
}
\end{center}
\end{table*}
\subsection{Main results}\label{subsec:main_results}

\textbf{Quantitative results.}
As demonstrated in Table~\ref{tab:main}, \sname~significantly enhances the performance of various baseline models. For instance, when applied to Gemini-2.0-Flash, \sname~improves AIDE's any medal achieving rates in Kaggle competitions from 25.8\% to 43.9\%, representing an improvement of over 18 percentage points, and rate of above median from 39.4\% to 63.6\%.
Notably, \sname~with Gemini-2.0-Flash also substantially outperforms AIDE using a powerful reasoning model (\ie, o1-preview) in terms of achieving gold medals in 10\% more tasks.
Moreoever, using Gemini-2.5-Pro, MLE-STAR shows a medal achievement rate of over 60\%.

\noindent\textbf{Comparison to DS-Agent.}
While DS-Agent~\citep{guo2024ds} shows competitive results on ML tasks, it necessitates human effort to curate its case bank from Kaggle.
Consequently, a direct comparison between DS-Agent and AIDE or our method is not feasible, as collecting tasks across diverse modalities, such as audio classification or image denoising, requires additional effort.
Nevertheless, we utilize four tabular classification tasks, \ie, wild-blueberry-yield (WBY), media-campaign-cost (MCC), spaceship-titanic (ST), and enzyme-substrate (ES), the same ones employed during DS-Agent's development stage~\citep{guo2024ds}, for a comparison. All experiments are done for 5 seeds following the original setup.
As shown in Table~\ref{tab:ds-agent}, \sname~using Gemini-2.0-Flash significantly outperforms DS-Agent even without human efforts.
See Appendix~\ref{app:extra_ds} for additional results, including comparison with AutoGluon~\citep{erickson2020autogluon}.

\subsection{Ablation studies}\label{subsec:ablations}

\textbf{Performance with reasoning models.}
First of all, as shown in Table~\ref{tab:main}, Gemini-2.5-Pro yields better performance than using Gemini-2.0-Flash. For example, in denoising-dirty-documents cometition, \sname~wigh Gemini-2.0-Flash scored above the median across all three seeds, failing to achieve any medals. However, when using Gemini-2.5-Pro, \sname~achieves two gold medals and one silver medal. These results demonstrate that \sname~is designed to harness the advancements of rapidly improving reasoning-based LLMs.

In addition, we conducted additional experiments using Claude-Sonnet-4. As shown in Table~\ref{tab:gemini-2.5}, other models besides Gemini also show promising results, proving compatibility and generalizability in terms of model types.
Here, we select four different type of competitions: image-to-image (denoising-dirty-documents; DDD), image classification (dog-breed-identification; DBI), text classification (spooky-author-identification, SAI), and audio classification (the-icml-2013-whale-challenge-right-whale-redux; WCR). We run each competition for three seeds. These results indicates that our framework is also compatible and generalizable in terms of LLM type.

\noindent\textbf{Effectiveness of proposed ensemble method.}
As highlighted in Table~\ref{tab:abl_tts}, \sname~demonstrates a significant performance improvement over the competing baseline, \ie, AIDE, achieving over a 12\% higher rate of obtaining any medal \textit{even without} additional ensemble strategy.
Notably, by ensembling multiple solution candidates, our approach yields even greater performance gains, \ie, \sname~consistently improves the success rate for achieving any medal (and specifically gold medals), also surpassing the median human expert's performance by a larger margin compared to scenarios where this ensembling method is not used.
While simpler strategies, such as selecting the solution with the best validation score or averaging final submissions, also offer benefits, \sname~shows stronger effectiveness, \eg, leading to a higher number of gold medals.

\begin{figure}[t]
\begin{minipage}[b]{0.48\textwidth} 
\centering
\includegraphics[width=\linewidth]{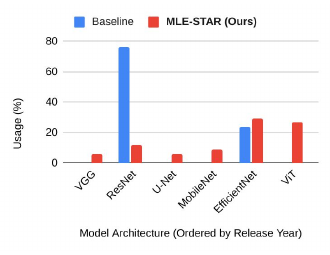}
\vspace{-0.25in}
\caption{\textbf{Model usage} (\%) on image classification competitions. Other models (11.7\%), which are used by \sname, are omitted.}
\label{fig:model_usage}
\end{minipage} \hfill 
\begin{minipage}[b]{0.48\textwidth} 
\centering
\includegraphics[width=0.95\linewidth]{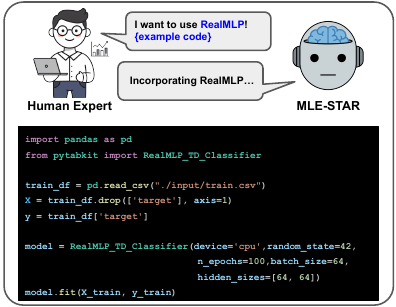}
\caption{\textbf{Human intervention.} By manually adding a model description, \sname~integrates its training into the framework.}
\label{fig:intervention}
\end{minipage}
\end{figure}

\section{Discussion}\label{sec:discussion}

\begin{figure}[t]
\begin{minipage}[b]{0.49\textwidth} 
\centering
\includegraphics[width=\linewidth]{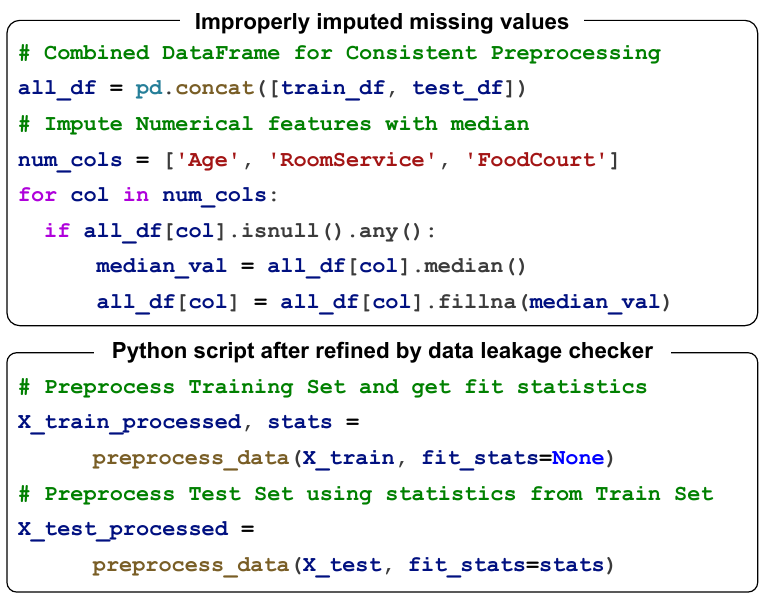}
\caption{\sname's \textbf{data leakage checker} introduces appropriate preprocessing.}
\label{fig:data_leakage_checker}
\end{minipage} \hfill 
\begin{minipage}[b]{0.49\textwidth} 
\centering
\includegraphics[width=\linewidth]{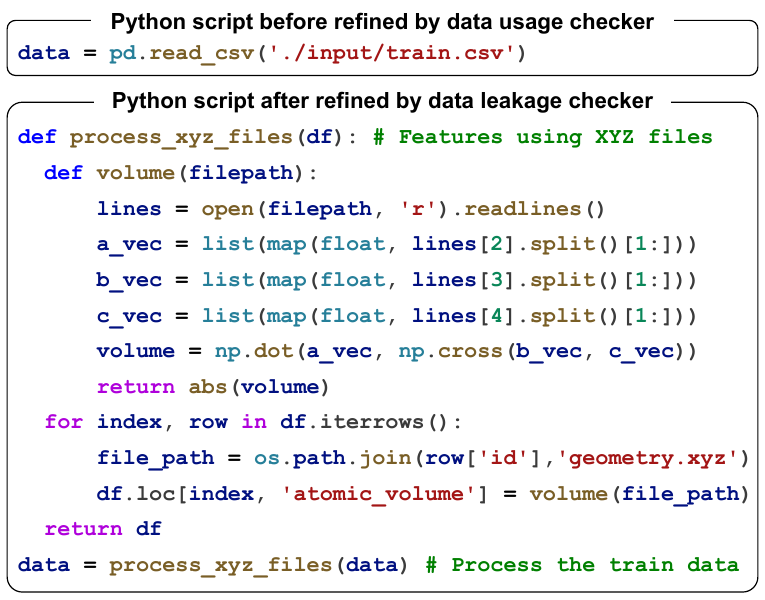}
\caption{\sname's \textbf{data usage checker} captures previously unused information.}
\label{fig:data_usage_checker}
\end{minipage}

\end{figure}


\textbf{Qualitative observations on selected models.} Figure~\ref{fig:model_usage} illustrates the model usage of two MLE agents: AIDE and \sname. AIDE primarily employs ResNet~\citep{he2016deep} for image classification. However, ResNet, released in 2015, is now considered outdated and can result in suboptimal performance.
In contrast, our \sname~primarily utilizes more recent and competitive models like EfficientNet~\citep{tan2019efficientnet} or ViT~\citep{dosovitskiy2021an}, leading to the performance gain, winning 37\% of the medals, more than AIDE, which wins 26\% of the image classification challenges.

\begin{table*}[t]
\begin{minipage}{0.56\textwidth}
\caption{Improvement failure when not using data leakage checker $\mathcal{A}_\mathtt{leakage}$ on spaceship-titanic competition.}\label{tab:st_failure}
\vspace{-0.3in}
\center\small
\begin{tabular}{lc}
\toprule
Metric & Accuracy ($\uparrow$)\\
\midrule
Validation & 0.8188 $\rightarrow$ \textbf{0.8677}\\
Test & \textbf{0.8033} $\rightarrow$ 0.7343\\
\bottomrule
\end{tabular}
\vspace{0.25in}
\caption{Ablation study of data usage checker $\mathcal{A}_\mathtt{data}$ on nomad2018-predicting competition.}\label{tab:nomad_ex}
\vspace{-0.3in}
\center\small
\begin{tabular}{lcc}
\toprule
Model & $\mathcal{A}_\mathtt{data}$ & RMSLE ($\downarrow$)\\
\midrule
\sname & \textcolor{red}{\ding{55}} & 0.0591\\
\textbf{\sname} & \textcolor{green}{\ding{51}} & \textbf{0.0559}\\
\bottomrule
\end{tabular}
\end{minipage} \hfill
\centering
\begin{minipage}{0.4\textwidth}
\centering
\includegraphics[width=0.99\linewidth]{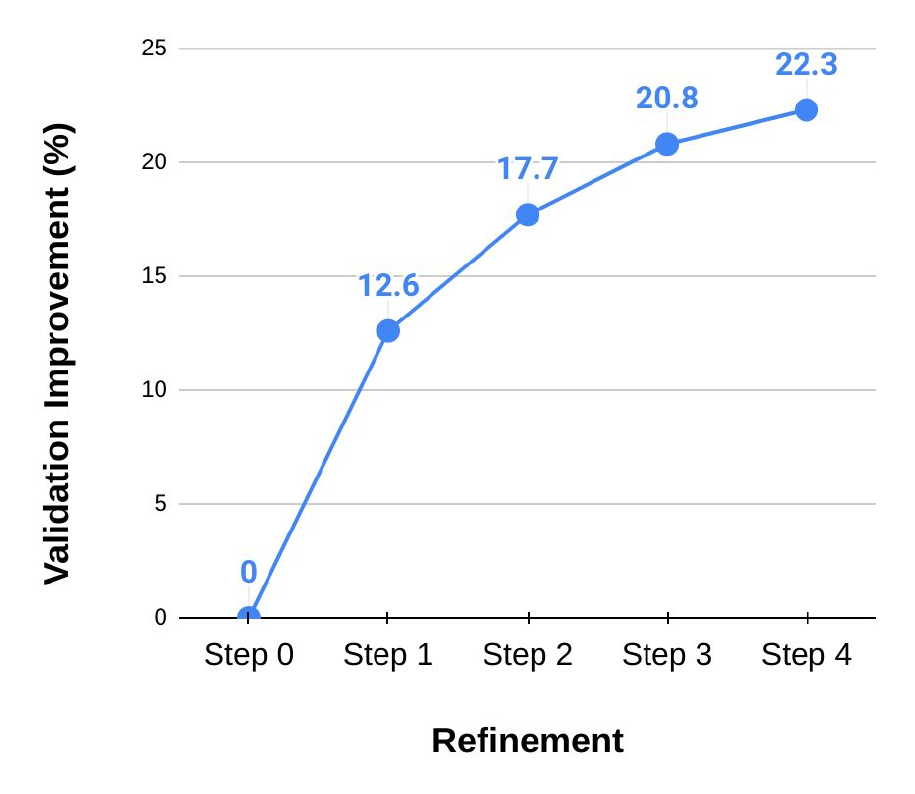}
\captionof{figure}{Solution refinement trajectory.}
\label{fig:improvement}
\end{minipage}
\end{table*}

\noindent\textbf{Human intervention.} \sname~readily adopts even more recent models with minimal human intervention. While \sname~automatically constructs a model description $\{\mathcal{T}_\mathtt{model}, \mathcal{T}_\mathtt{code}\}$ using search as tool, a natural extension involves leveraging human expertise for this construction. As shown in Figure~\ref{fig:intervention}, by manually adding a model description for RealMLP~\citep{holzmuller2024better}, \sname~successfully integrates its training into the framework, a model not previously retrieved. 
In addition, users can also specify the target code blocks by replacing the ablation summary with manually written instructions.


\noindent\textbf{Misbehavior of LLMs and corrections.}
We observe that while the code generated by the LLM executed correctly, their content is sometime unrealistic, exhibiting hallucination.
For example, Figure~\ref{fig:data_leakage_checker} illustrates an impractical approach where test data is preprocessed using its own statistics. Since test data must remain unseen, correction in the code is necessitated, for which, \sname~employs a data leakage checker $\mathcal{A}_\mathtt{leakage}$ to identify such issues in the generated Python script. If a problem is detected, \sname~refines the code.
As shown in the Figure, \sname~successfully identifies the issue and modifies the code by, first extracting statistics from the training data and then preprocessing the test data using these calculated statistics.
In addition, the improvement process can fail to generalize when $\mathcal{A}_\mathtt{leakage}$ is not employed, as exemplified in Table~\ref{tab:st_failure}. In this example, the validation accuracy (\ie, the target objective) improves, but the test accuracy drops significantly. This is attributed to the LLM performing feature engineering using the target variable \texttt{Transported}, which is not accessible in the test set, leading to data leakage and subsequently, poor test performance.

We also observe that LLMs often generate Python scripts that overlook some of the provided data sources. For example, in the nomad2018-predicting competition, Gemini-2.0-Flash solely loads train.csv, neglecting the use of geometry.xyz (see Figure~\ref{fig:data_usage_checker}).
To address this, \sname~employs $\mathcal{A}_\mathtt{data}$, which reexamines the task description to ensure that all given data is utilized. As shown in Figure~\ref{fig:data_usage_checker}, this design enables \sname~to incorporate previously neglected data. As a result, performance is significantly improved, as shown in Table~\ref{tab:nomad_ex}.

\noindent\textbf{Progressive improvement via \sname~refinement.}
This section details the progressive improvement of solutions achieved by \sname, as measured by validation metrics. Given the task-specific nature of evaluation metrics, we report the average relative error reduction (\%) across the all 22 challenges in MLE-bench Lite~\citep{chan2025mle}. This metric measures the extent to which \sname~reduces the error of an initial solution.
Figure~\ref{fig:improvement} demonstrates a consistent improvement as \sname~proceeds through its refinement steps, which each step focusing on refining a single code block via an inner loop. Significantly, the magnitude of improvement is notable in the early refinement stages. We posit that this stems from \sname's ablation study module which helps to target the most influential code blocks for modification first.
\section{Conclusion}\label{sec:conclusion}
We propose \sname, a novel MLE agent designed for various ML tasks.
Our key idea is to utilize a search engine to retrieve effective models and then explore various strategies targeting specific ML pipeline components to improve the solution.
The effectiveness of \sname~is validated by winning medals in 64\% (where 36\% are gold medals) of the MLE-bench Lite Kaggle competitions.

\noindent\textbf{Limitation.}
Since Kaggle competitions are publicly accessible, there is a potential risk that LLMs might have been trained with the relevant discussions about the challenge. Nevertheless, we show that \sname's solution is sufficiently novel (using LLM as a judge) compared to the discussions on Kaggle (see Appendix~\ref{app:contamination}).

\bibliographystyle{abbrvnat}
\nobibliography*
\bibliography{main}

\begin{thebibliography}{50}
\providecommand{\natexlab}[1]{#1}
\providecommand{\url}[1]{\texttt{#1}}
\expandafter\ifx\csname urlstyle\endcsname\relax
  \providecommand{\doi}[1]{doi: #1}\else
  \providecommand{\doi}{doi: \begingroup \urlstyle{rm}\Url}\fi

\bibitem[Brown et~al.(2020)Brown, Mann, Ryder, Subbiah, Kaplan, Dhariwal, Neelakantan, Shyam, Sastry, Askell, et~al.]{brown2020language}
T.~Brown, B.~Mann, N.~Ryder, M.~Subbiah, J.~D. Kaplan, P.~Dhariwal, A.~Neelakantan, P.~Shyam, G.~Sastry, A.~Askell, et~al.
\newblock Language models are few-shot learners.
\newblock \emph{Advances in Neural Information Processing Systems}, 2020.

\bibitem[Chan et~al.(2025)Chan, Chowdhury, Jaffe, Aung, Sherburn, Mays, Starace, Liu, Maksin, Patwardhan, et~al.]{chan2025mle}
J.~S. Chan, N.~Chowdhury, O.~Jaffe, J.~Aung, D.~Sherburn, E.~Mays, G.~Starace, K.~Liu, L.~Maksin, T.~Patwardhan, et~al.
\newblock Mle-bench: Evaluating machine learning agents on machine learning engineering.
\newblock \emph{International Conference on Learning Representations}, 2025.

\bibitem[Chen and Guestrin(2016)]{chen2016xgboost}
T.~Chen and C.~Guestrin.
\newblock Xgboost: A scalable tree boosting system.
\newblock \emph{ACM SIGKDD International Conference on Knowledge Discovery and Data Mining}, 2016.

\bibitem[Dosovitskiy et~al.(2021)Dosovitskiy, Beyer, Kolesnikov, Weissenborn, Zhai, Unterthiner, Dehghani, Minderer, Heigold, Gelly, Uszkoreit, and Houlsby]{dosovitskiy2021an}
A.~Dosovitskiy, L.~Beyer, A.~Kolesnikov, D.~Weissenborn, X.~Zhai, T.~Unterthiner, M.~Dehghani, M.~Minderer, G.~Heigold, S.~Gelly, J.~Uszkoreit, and N.~Houlsby.
\newblock An image is worth 16x16 words: Transformers for image recognition at scale.
\newblock \emph{International Conference on Learning Representations}, 2021.

\bibitem[Elsken et~al.(2019)Elsken, Metzen, and Hutter]{elsken2019neural}
T.~Elsken, J.~H. Metzen, and F.~Hutter.
\newblock Neural architecture search: A survey.
\newblock \emph{Journal of Machine Learning Research}, 2019.

\bibitem[Erickson et~al.(2020)Erickson, Mueller, Shirkov, Zhang, Larroy, Li, and Smola]{erickson2020autogluon}
N.~Erickson, J.~Mueller, A.~Shirkov, H.~Zhang, P.~Larroy, M.~Li, and A.~Smola.
\newblock Autogluon-tabular: Robust and accurate automl for structured data.
\newblock \emph{arXiv preprint arXiv:2003.06505}, 2020.

\bibitem[Fan et~al.(2019)Fan, Zhang, Fan, and Zhang]{fan2019brief}
L.~Fan, F.~Zhang, H.~Fan, and C.~Zhang.
\newblock Brief review of image denoising techniques.
\newblock \emph{Visual computing for industry, biomedicine, and art}, 2019.

\bibitem[Fan et~al.(2010)Fan, Zhong, Peng, Verscheure, Zhang, Ren, Yan, and Yang]{fan2010generalized}
W.~Fan, E.~Zhong, J.~Peng, O.~Verscheure, K.~Zhang, J.~Ren, R.~Yan, and Q.~Yang.
\newblock Generalized and heuristic-free feature construction for improved accuracy.
\newblock \emph{SIAM International Conference on Data Mining}, 2010.

\bibitem[Feurer et~al.(2022)Feurer, Eggensperger, Falkner, Lindauer, and Hutter]{feurer2022auto}
M.~Feurer, K.~Eggensperger, S.~Falkner, M.~Lindauer, and F.~Hutter.
\newblock Auto-sklearn 2.0: Hands-free automl via meta-learning.
\newblock \emph{Journal of Machine Learning Research}, 2022.

\bibitem[Guo et~al.(2024)Guo, Deng, Wen, Chen, Chang, and Wang]{guo2024ds}
S.~Guo, C.~Deng, Y.~Wen, H.~Chen, Y.~Chang, and J.~Wang.
\newblock {DS}-agent: Automated data science by empowering large language models with case-based reasoning.
\newblock \emph{International Conference on Machine Learning}, 2024.

\bibitem[He et~al.(2016)He, Zhang, Ren, and Sun]{he2016deep}
K.~He, X.~Zhang, S.~Ren, and J.~Sun.
\newblock Deep residual learning for image recognition.
\newblock \emph{IEEE Conference on Computer Vision and Pattern Recognition}, 2016.

\bibitem[Hollmann et~al.(2023)Hollmann, M{\"u}ller, and Hutter]{hollmann2023large}
N.~Hollmann, S.~M{\"u}ller, and F.~Hutter.
\newblock Large language models for automated data science: Introducing caafe for context-aware automated feature engineering.
\newblock \emph{Advances in Neural Information Processing Systems}, 2023.

\bibitem[Hollmann et~al.(2025)Hollmann, M{\"u}ller, Purucker, Krishnakumar, K{\"o}rfer, Hoo, Schirrmeister, and Hutter]{hollmann2025accurate}
N.~Hollmann, S.~M{\"u}ller, L.~Purucker, A.~Krishnakumar, M.~K{\"o}rfer, S.~B. Hoo, R.~T. Schirrmeister, and F.~Hutter.
\newblock Accurate predictions on small data with a tabular foundation model.
\newblock \emph{Nature}, 2025.

\bibitem[Holzm{\"u}ller et~al.(2024)Holzm{\"u}ller, Grinsztajn, and Steinwart]{holzmuller2024better}
D.~Holzm{\"u}ller, L.~Grinsztajn, and I.~Steinwart.
\newblock Better by default: Strong pre-tuned mlps and boosted trees on tabular data.
\newblock \emph{Advances in Neural Information Processing Systems}, 2024.

\bibitem[Hong et~al.(2024)Hong, Lin, Liu, Liu, Wu, Zhang, Wei, Li, Chen, Zhang, et~al.]{hong2024data}
S.~Hong, Y.~Lin, B.~Liu, B.~Liu, B.~Wu, C.~Zhang, C.~Wei, D.~Li, J.~Chen, J.~Zhang, et~al.
\newblock Data interpreter: An llm agent for data science.
\newblock \emph{arXiv preprint arXiv:2402.18679}, 2024.

\bibitem[Horn et~al.(2019)Horn, Pack, and Rieger]{horn2019autofeat}
F.~Horn, R.~Pack, and M.~Rieger.
\newblock The autofeat python library for automated feature engineering and selection.
\newblock \emph{Joint European Conference on Machine Learning and Knowledge Discovery in Databases}, 2019.

\bibitem[Hu et~al.(2024)Hu, Zhao, Wei, Chai, Ma, Wang, Wang, Su, Xu, Zhu, et~al.]{hu2024infiagent}
X.~Hu, Z.~Zhao, S.~Wei, Z.~Chai, Q.~Ma, G.~Wang, X.~Wang, J.~Su, J.~Xu, M.~Zhu, et~al.
\newblock Infiagent-dabench: Evaluating agents on data analysis tasks.
\newblock \emph{arXiv preprint arXiv:2401.05507}, 2024.

\bibitem[Huang et~al.(2024{\natexlab{a}})Huang, Vora, Liang, and Leskovec]{huang2024mlagentbench}
Q.~Huang, J.~Vora, P.~Liang, and J.~Leskovec.
\newblock Mlagentbench: Evaluating language agents on machine learning experimentation.
\newblock \emph{International Conference on Machine Learning}, 2024{\natexlab{a}}.

\bibitem[Huang et~al.(2024{\natexlab{b}})Huang, Luo, Yu, Zhang, Lei, Wei, He, Huang, Liu, Zhao, et~al.]{huang2024code}
Y.~Huang, J.~Luo, Y.~Yu, Y.~Zhang, F.~Lei, Y.~Wei, S.~He, L.~Huang, X.~Liu, J.~Zhao, et~al.
\newblock Da-code: Agent data science code generation benchmark for large language models.
\newblock \emph{arXiv preprint arXiv:2410.07331}, 2024{\natexlab{b}}.

\bibitem[Ichihara et~al.(2025)Ichihara, Jinnai, Morimura, Abe, Ariu, Sakamoto, and Uchibe]{ichihara2025evaluation}
Y.~Ichihara, Y.~Jinnai, T.~Morimura, K.~Abe, K.~Ariu, M.~Sakamoto, and E.~Uchibe.
\newblock Evaluation of best-of-n sampling strategies for language model alignment.
\newblock \emph{Transactions on Machine Learning Research}, 2025.

\bibitem[Jain et~al.(2025)Jain, Han, Gu, Li, Yan, Zhang, Wang, Solar-Lezama, Sen, and Stoica]{jain2025livecodebench}
N.~Jain, K.~Han, A.~Gu, W.-D. Li, F.~Yan, T.~Zhang, S.~Wang, A.~Solar-Lezama, K.~Sen, and I.~Stoica.
\newblock Livecodebench: Holistic and contamination free evaluation of large language models for code.
\newblock \emph{International Conference on Learning Representations}, 2025.

\bibitem[Jiang et~al.(2025)Jiang, Schmidt, Srikanth, Xu, Kaplan, Jacenko, and Wu]{jiang2025aide}
Z.~Jiang, D.~Schmidt, D.~Srikanth, D.~Xu, I.~Kaplan, D.~Jacenko, and Y.~Wu.
\newblock Aide: Ai-driven exploration in the space of code.
\newblock \emph{arXiv preprint arXiv:2502.13138}, 2025.

\bibitem[Jimenez et~al.(2024)Jimenez, Yang, Wettig, Yao, Pei, Press, and Narasimhan]{jimenez2024swe}
C.~E. Jimenez, J.~Yang, A.~Wettig, S.~Yao, K.~Pei, O.~Press, and K.~Narasimhan.
\newblock Swe-bench: Can language models resolve real-world github issues?
\newblock \emph{International Conference on Learning Representations}, 2024.

\bibitem[Jin et~al.(2019)Jin, Song, and Hu]{jin2019auto}
H.~Jin, Q.~Song, and X.~Hu.
\newblock Auto-keras: An efficient neural architecture search system.
\newblock \emph{ACM SIGKDD International Conference on Knowledge Discovery and Data Mining}, 2019.

\bibitem[Jing et~al.(2025)Jing, Huang, Wang, Yao, Yu, Ma, Zhang, Du, and Yu]{jing2024dsbench}
L.~Jing, Z.~Huang, X.~Wang, W.~Yao, W.~Yu, K.~Ma, H.~Zhang, X.~Du, and D.~Yu.
\newblock Dsbench: How far are data science agents to becoming data science experts?
\newblock \emph{International Conference on Learning Representations}, 2025.

\bibitem[Kanter and Veeramachaneni(2015)]{kanter2015deep}
J.~M. Kanter and K.~Veeramachaneni.
\newblock Deep feature synthesis: Towards automating data science endeavors.
\newblock \emph{IEEE International Conference on Data Science and Advanced Analytics}, 2015.

\bibitem[Kolodner(1992)]{kolodner1992introduction}
J.~L. Kolodner.
\newblock An introduction to case-based reasoning.
\newblock \emph{Artificial intelligence review}, 1992.

\bibitem[Kotthoff et~al.(2017)Kotthoff, Thornton, Hoos, Hutter, and Leyton-Brown]{kotthoff2017auto}
L.~Kotthoff, C.~Thornton, H.~H. Hoos, F.~Hutter, and K.~Leyton-Brown.
\newblock Auto-weka 2.0: Automatic model selection and hyperparameter optimization in weka.
\newblock \emph{Journal of Machine Learning Research}, 2017.

\bibitem[LeDell and Poirier(2020)]{H2OAutoML20}
E.~LeDell and S.~Poirier.
\newblock {H2O} {A}uto{ML}: Scalable automatic machine learning.
\newblock \emph{ICML Workshop on AutoML}, 2020.

\bibitem[Li et~al.(2023)Li, Wang, Zha, Huang, Wu, Chen, and Zhao]{li2023learning}
L.~Li, H.~Wang, L.~Zha, Q.~Huang, S.~Wu, G.~Chen, and J.~Zhao.
\newblock Learning a data-driven policy network for pre-training automated feature engineering.
\newblock \emph{International Conference on Learning Representations}, 2023.

\bibitem[Li et~al.(2022)Li, Choi, Chung, Kushman, Schrittwieser, Leblond, Eccles, Keeling, Gimeno, Dal~Lago, et~al.]{li2022competition}
Y.~Li, D.~Choi, J.~Chung, N.~Kushman, J.~Schrittwieser, R.~Leblond, T.~Eccles, J.~Keeling, F.~Gimeno, A.~Dal~Lago, et~al.
\newblock Competition-level code generation with alphacode.
\newblock \emph{Science}, 2022.

\bibitem[Li et~al.(2024)Li, Zang, Ma, Guo, Zheng, Liu, Niu, Wang, Yang, Liu, et~al.]{li2024autokaggle}
Z.~Li, Q.~Zang, D.~Ma, J.~Guo, T.~Zheng, M.~Liu, X.~Niu, Y.~Wang, J.~Yang, J.~Liu, et~al.
\newblock Autokaggle: A multi-agent framework for autonomous data science competitions.
\newblock \emph{arXiv preprint arXiv:2410.20424}, 2024.

\bibitem[Nam et~al.(2024)Nam, Kim, Oh, Tack, Kim, and Shin]{nam2024optimized}
J.~Nam, K.~Kim, S.~Oh, J.~Tack, J.~Kim, and J.~Shin.
\newblock Optimized feature generation for tabular data via llms with decision tree reasoning.
\newblock \emph{Advances in Neural Information Processing Systems}, 2024.

\bibitem[Olson and Moore(2016)]{olson2016tpot}
R.~S. Olson and J.~H. Moore.
\newblock Tpot: A tree-based pipeline optimization tool for automating machine learning.
\newblock \emph{ICML Workshop on AutoML}, 2016.

\bibitem[Pedregosa et~al.(2011)Pedregosa, Varoquaux, Gramfort, Michel, Thirion, Grisel, Blondel, Prettenhofer, Weiss, Dubourg, et~al.]{pedregosa2011scikit}
F.~Pedregosa, G.~Varoquaux, A.~Gramfort, V.~Michel, B.~Thirion, O.~Grisel, M.~Blondel, P.~Prettenhofer, R.~Weiss, V.~Dubourg, et~al.
\newblock Scikit-learn: Machine learning in python.
\newblock \emph{Journal of Machine Learning Research}, 2011.

\bibitem[Pham et~al.(2018)Pham, Guan, Zoph, Le, and Dean]{pham2018efficient}
H.~Pham, M.~Guan, B.~Zoph, Q.~Le, and J.~Dean.
\newblock Efficient neural architecture search via parameters sharing.
\newblock \emph{International Conference on Machine Learning}, 2018.

\bibitem[Prokhorenkova et~al.(2018)Prokhorenkova, Gusev, Vorobev, Dorogush, and Gulin]{prokhorenkova2018catboost}
L.~Prokhorenkova, G.~Gusev, A.~Vorobev, A.~V. Dorogush, and A.~Gulin.
\newblock Catboost: unbiased boosting with categorical features.
\newblock \emph{Advances in Neural Information Processing Systems}, 2018.

\bibitem[Real et~al.(2019)Real, Aggarwal, Huang, and Le]{real2019regularized}
E.~Real, A.~Aggarwal, Y.~Huang, and Q.~V. Le.
\newblock Regularized evolution for image classifier architecture search.
\newblock \emph{AAAI Conference on Artificial Intelligence}, 2019.

\bibitem[Schmidgall et~al.(2025)Schmidgall, Su, Wang, Sun, Wu, Yu, Liu, Liu, and Barsoum]{schmidgall2025agent}
S.~Schmidgall, Y.~Su, Z.~Wang, X.~Sun, J.~Wu, X.~Yu, J.~Liu, Z.~Liu, and E.~Barsoum.
\newblock Agent laboratory: Using llm agents as research assistants.
\newblock \emph{arXiv preprint arXiv:2501.04227}, 2025.

\bibitem[Shen et~al.(2023)Shen, Song, Tan, Li, Lu, and Zhuang]{shen2023hugginggpt}
Y.~Shen, K.~Song, X.~Tan, D.~Li, W.~Lu, and Y.~Zhuang.
\newblock Hugginggpt: Solving ai tasks with chatgpt and its friends in hugging face.
\newblock \emph{Advances in Neural Information Processing Systems}, 2023.

\bibitem[Tan and Le(2019)]{tan2019efficientnet}
M.~Tan and Q.~Le.
\newblock Efficientnet: Rethinking model scaling for convolutional neural networks.
\newblock \emph{International Conference on Machine Learning}, 2019.

\bibitem[Team et~al.(2024)Team, Georgiev, Lei, Burnell, Bai, Gulati, Tanzer, Vincent, Pan, Wang, et~al.]{team2024gemini}
G.~Team, P.~Georgiev, V.~I. Lei, R.~Burnell, L.~Bai, A.~Gulati, G.~Tanzer, D.~Vincent, Z.~Pan, S.~Wang, et~al.
\newblock Gemini 1.5: Unlocking multimodal understanding across millions of tokens of context.
\newblock \emph{arXiv preprint arXiv:2403.05530}, 2024.

\bibitem[Touvron et~al.(2023)Touvron, Lavril, Izacard, Martinet, Lachaux, Lacroix, Rozi{\`e}re, Goyal, Hambro, Azhar, et~al.]{touvron2023llama}
H.~Touvron, T.~Lavril, G.~Izacard, X.~Martinet, M.-A. Lachaux, T.~Lacroix, B.~Rozi{\`e}re, N.~Goyal, E.~Hambro, F.~Azhar, et~al.
\newblock Llama: Open and efficient foundation language models.
\newblock \emph{arXiv preprint arXiv:2302.13971}, 2023.

\bibitem[Wang et~al.(2023)Wang, Xie, Jiang, Mandlekar, Xiao, Zhu, Fan, and Anandkumar]{wang2023voyager}
G.~Wang, Y.~Xie, Y.~Jiang, A.~Mandlekar, C.~Xiao, Y.~Zhu, L.~Fan, and A.~Anandkumar.
\newblock Voyager: An open-ended embodied agent with large language models.
\newblock \emph{arXiv preprint arXiv: Arxiv-2305.16291}, 2023.

\bibitem[Wang et~al.(2024)Wang, Li, Song, Xu, Tang, Zhuge, Pan, Song, Li, Singh, et~al.]{wang2024openhands}
X.~Wang, B.~Li, Y.~Song, F.~F. Xu, X.~Tang, M.~Zhuge, J.~Pan, Y.~Song, B.~Li, J.~Singh, et~al.
\newblock Openhands: An open platform for ai software developers as generalist agents.
\newblock \emph{International Conference on Learning Representations}, 2024.

\bibitem[Watson and Marir(1994)]{watson1994case}
I.~Watson and F.~Marir.
\newblock Case-based reasoning: A review.
\newblock \emph{The knowledge engineering review}, 1994.

\bibitem[Yao et~al.(2023)Yao, Zhao, Yu, Du, Shafran, Narasimhan, and Cao]{yao2023react}
S.~Yao, J.~Zhao, D.~Yu, N.~Du, I.~Shafran, K.~Narasimhan, and Y.~Cao.
\newblock React: Synergizing reasoning and acting in language models.
\newblock \emph{International Conference on Learning Representations}, 2023.

\bibitem[You et~al.(2025)You, Zhang, Xu, Lou, Yan, Wang, Zhang, and Huang]{you2025datawiseagent}
Z.~You, Y.~Zhang, D.~Xu, Y.~Lou, Y.~Yan, W.~Wang, H.~Zhang, and Y.~Huang.
\newblock Datawiseagent: A notebook-centric llm agent framework for automated data science.
\newblock \emph{arXiv preprint arXiv:2503.07044}, 2025.

\bibitem[Zhang et~al.(2023)Zhang, Zhang, Fan, Luo, Liu, Liu, Cao, and Jian]{zhang2023openfe}
T.~Zhang, Z.~A. Zhang, Z.~Fan, H.~Luo, F.~Liu, Q.~Liu, W.~Cao, and L.~Jian.
\newblock Openfe: Automated feature generation with expert-level performance.
\newblock \emph{International Conference on Machine Learning}, 2023.

\bibitem[Zoph and Le(2017)]{zoph2016neural}
B.~Zoph and Q.~V. Le.
\newblock Neural architecture search with reinforcement learning.
\newblock \emph{International Conference on Learning Representations}, 2017.

\end{thebibliography}

\clearpage

\newpage
\appendix
\begin{center}
{\bf {\Large Appendix}} 
\end{center}

\section{Prompts for \sname}\label{app:prompts}

\subsection{Retriever agent}
\begin{figure*}[h]
\centering
\includegraphics[width=0.85\textwidth]{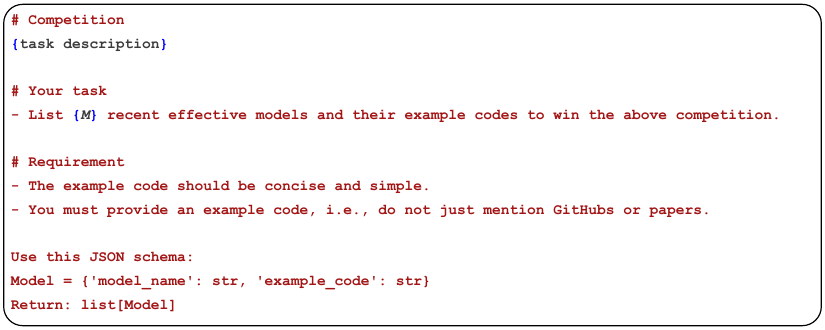}
\caption{
Prompt used for retrieving task-specific models using web search.
}
\label{prompt:retriever}
\end{figure*}



\sname~starts by generating an initial solution.
Here, we propose using web search as a tool for \sname~first to retrieve $M$ state-of-the-art models for the given task.
Specifically, \sname~leverages a retriever agent $\mathcal{A}_\mathtt{retriever}$ with the above prompt (Figure~\ref{prompt:retriever}).
$\mathcal{A}_\mathtt{retriever}$ takes task description $\mathcal{T}_\mathtt{task}$ as input and retrieves $M$ pairs of $\{\mathcal{T}_\mathtt{model},\mathcal{T}_\mathtt{code}\}$.
Here, we guide \sname~to generate the retrieved result as structured output (\ie, JSON).
After we obtain JSON file, we parse them into separate model cards.\footnote{See $\texttt{example\_intermediate\_outputs/retriever\_output.txt}$ in \url{https://github.com/jaehyun513/MLE-STAR}.}

\newpage
\subsection{Candidate evaluation agent}
\begin{figure*}[h]
\centering
\includegraphics[width=0.85\textwidth]{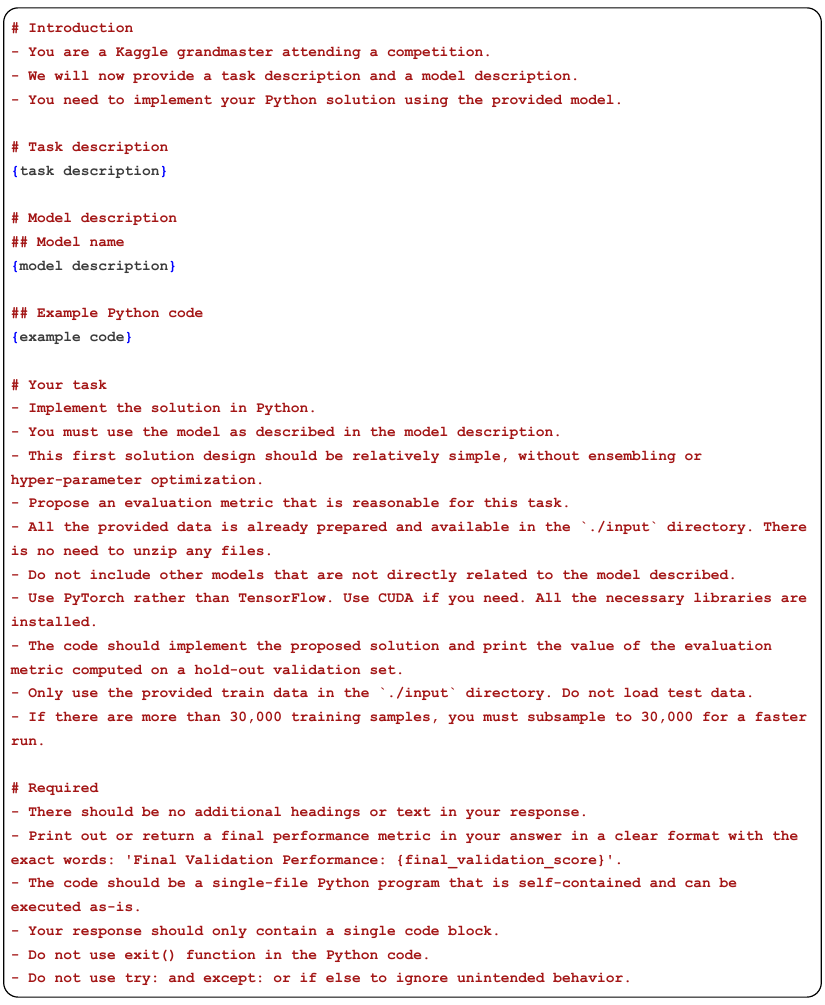}
\caption{
Prompt used for evaluating retrieved models.
}
\label{prompt:init}
\end{figure*}
\sname~uses candidate evaluation agent $\mathcal{A}_\mathtt{init}$ to evaluate the performance of the retrieved model.
As shown in Figure~\ref{prompt:init}, by taking task description ($\mathcal{T}_\mathtt{task}$), model description ($\mathcal{T}_\mathtt{model}$), and corresponding code example ($\mathcal{T}_\mathtt{code}$), $\mathcal{A}_\mathtt{init}$ generates a Python script.\footnote{See $\texttt{example\_intermediate\_outputs/candidate\_evaluation.py}$ in \url{https://github.com/jaehyun513/MLE-STAR} for an example.}
The Python script for the retrieved model evaluation is guided to be relatively simple, and to contain the evaluation result computed on a hold-out validation set.
In addition, if there are too many training samples, $\mathcal{A}_\mathtt{init}$ uses the subset of training sample for faster execution.

\newpage
\subsection{Merging agent}
\begin{figure*}[h]
\centering
\includegraphics[width=0.85\textwidth]{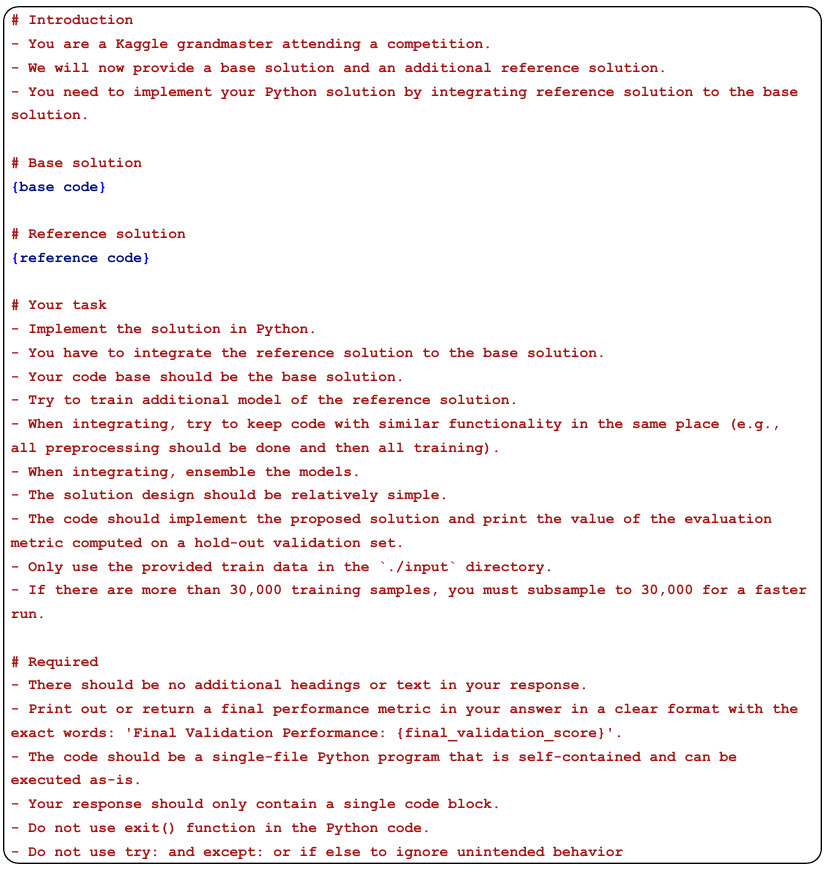}
\caption{
Prompt used for merging the candidate models for generating initial solution.
}
\label{prompt:merger}
\end{figure*}
\sname~leverages an agent $\mathcal{A}_\mathtt{merger}$ to merge the retrieved models into a consolidated initial solution.
As shown in Figure~\ref{prompt:merger}, this process is done sequentially, where the prompt guides the agent to integrate the reference code (\ie, the best candidate model code among the models that are not merged yet) into the base code (\ie, the current candidate merged script). The output of $\mathcal{A}_\mathtt{merger}$ is a Python script\footnote{See $\texttt{example\_intermediate\_outputs/merged\_candidate.py}$ in \url{https://github.com/jaehyun513/MLE-STAR} for an example.},
which will be the next candidate merged script. See Appendix~\ref{app:algorithm} for the sequential procedure of \sname~when generating the initial solution.

\newpage
\subsection{Ablation study agent}
\begin{figure*}[h]
\centering
\includegraphics[width=0.85\textwidth]{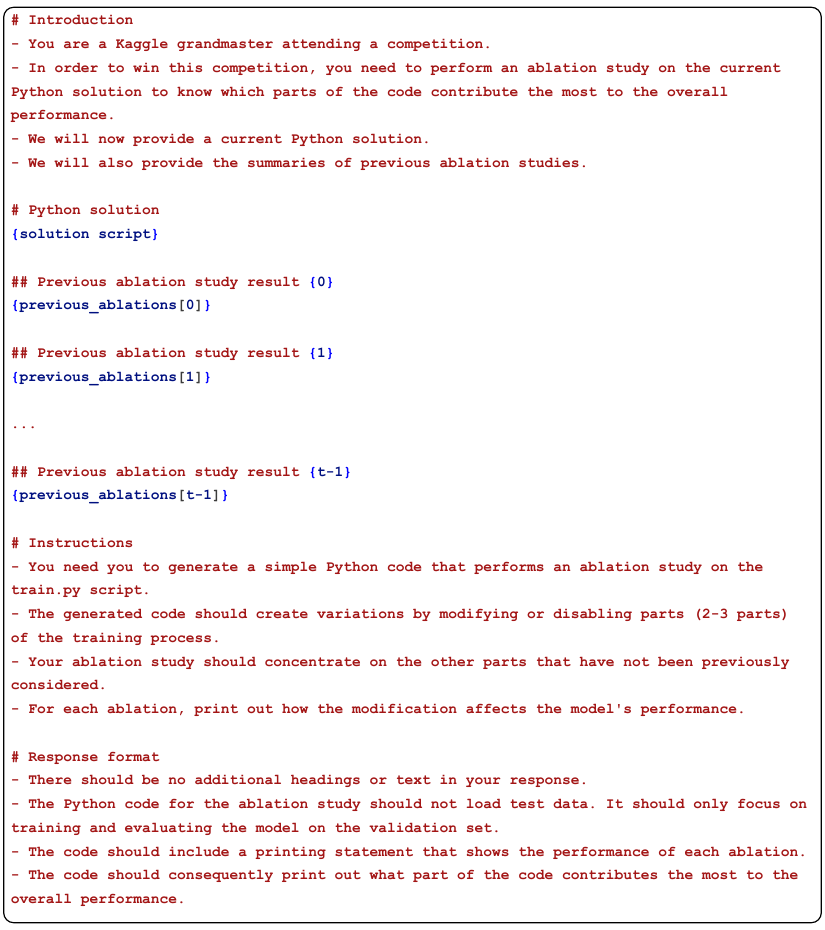}
\caption{
Prompt used for generating a Python script for ablation studies.
}
\label{prompt:abl}
\end{figure*}
To effectively explore specialized improvement strategies, \sname~identifies and targets specific code blocks. This code block selection is guided by an ablation study performed by an agent $\mathcal{A}_\mathtt{abl}$.
As shown in Figure~\ref{prompt:abl}, $\mathcal{A}_\mathtt{abl}$ generates a Python code designed to perform an ablation study on current solution. The prompt guides the agent to modify or disable specific component.\footnote{We provide an example generated code for ablation study (which is generated by $\mathcal{A}_\mathtt{abl}$) in \url{https://github.com/jaehyun513/MLE-STAR} (see $\texttt{example\_intermediate\_outputs/ablation.py}$).}
Moreover, to encourage exploration of different pipeline parts, the summaries of previous ablation studies are also used as valuable feedback. See Appendix~\ref{app:qual_ex} for the example output of the agent $\mathcal{A}_\mathtt{abl}$.

\newpage
\subsection{Ablation study summarization agent}
\begin{figure*}[h]
\centering
\includegraphics[width=0.85\textwidth]{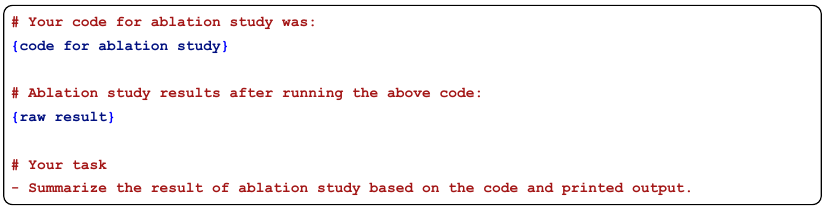}
\caption{
Prompt used for summarizing the result of the ablation study.
}
\label{prompt:summarize}
\end{figure*}
After executing the code for an ablation study, denoted as $a_t$, the output result $r_t$ is produced. Since $r_t$ often contains content unrelated to the ablation (for example, printing the loss value across training epochs), a summarization module $\mathcal{A}_\mathtt{summarize}$ is utilized with the prompt mentioned above (Figure~\ref{prompt:summarize}). This module takes $a_t$ and $r_t$ as input to summarize and parse the ablation study results. Here, $a_t$ is also used because it provides information about the modification. See Appendix~\ref{app:qual_ex} for the examples of $r_t$ and the summarization result.

\newpage
\subsection{Extractor}
\begin{figure*}[h]
\centering
\includegraphics[width=0.75\textwidth]{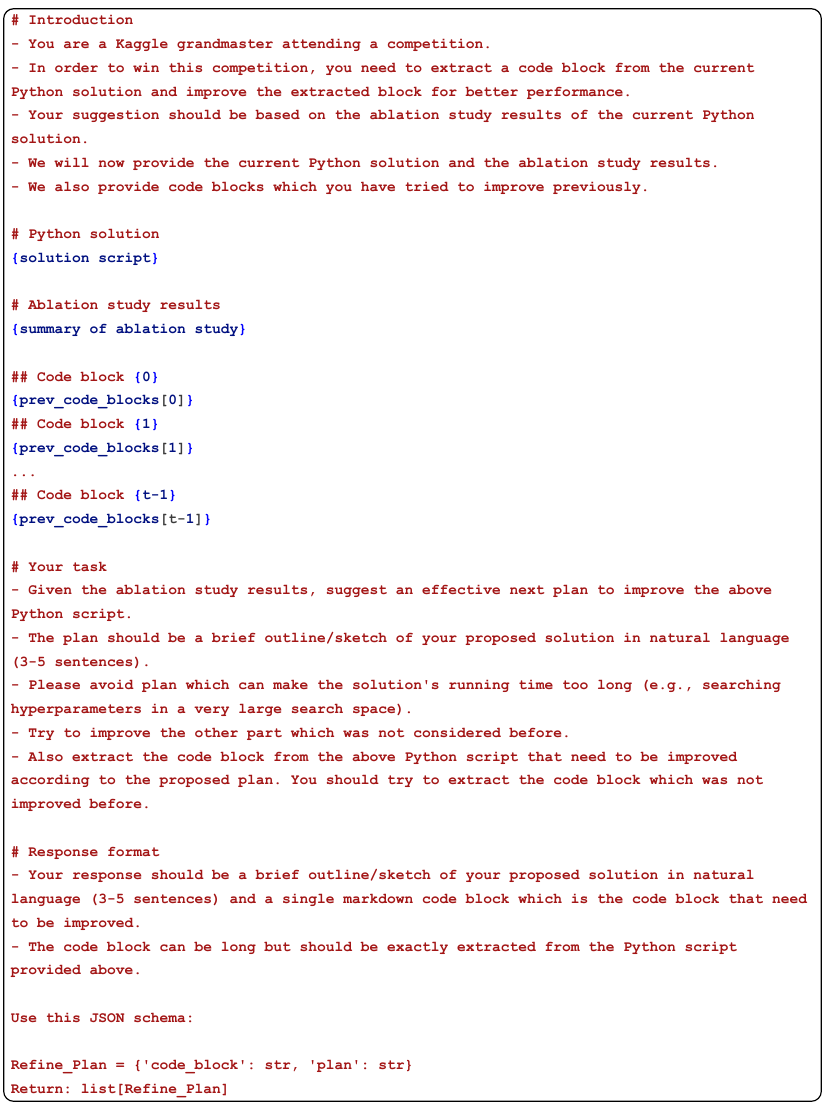}
\caption{
Prompt used for extracting the code block that has the most significant impact.
}
\label{prompt:extractor}
\end{figure*}
\sname~uses an extractor module $\mathcal{A}_\mathtt{extractor}$ to analyze the $\mathcal{T}_\mathtt{abl}$ and then identify the code block $c_t$. As shown in Figure~\ref{prompt:extractor}, $\mathcal{A}_\mathtt{extractor}$ takes the summary of the ablation study, current solution code, and the previously refined code blocks as input, and is guided to output the code block which has the most significant impact on performance. Here, the initial plan for refining the extracted code block is also generated.\footnote{See $\texttt{example\_intermediate\_outputs/code\_block.txt}$ in \url{https://github.com/jaehyun513/MLE-STAR} for an example of extracted code block.}

\newpage
\subsection{Coder}
\begin{figure*}[h]
\centering
\includegraphics[width=0.85\textwidth]{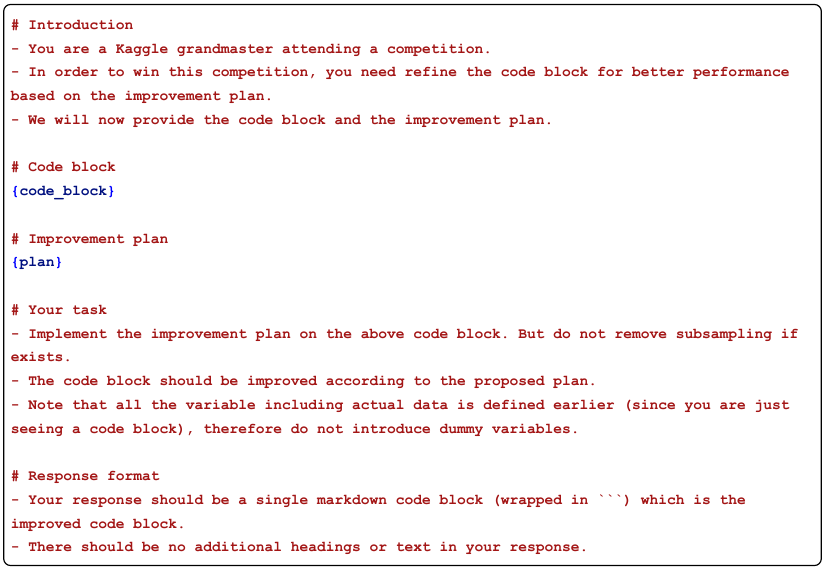}
\caption{
Prompt used for implementing refinement plan on the extracted code block.
}
\label{prompt:coder}
\end{figure*}
The implementation of code block refinement is done by $\mathcal{A}_\mathtt{coder}$, which takes the extracted code block and the refinement plan as input, and outputs the refined code block.\footnote{We provide an example of the target code block, proposed plan, and the output of refined code block by $\mathcal{A}_\mathtt{coder}$ in \url{https://github.com/jaehyun513/MLE-STAR} (see $\mathtt{example\_intermeidate\_outputs/coder\_outputs/}$ directory).}

\newpage
\subsection{Planner}
\begin{figure*}[h]
\centering
\includegraphics[width=0.85\textwidth]{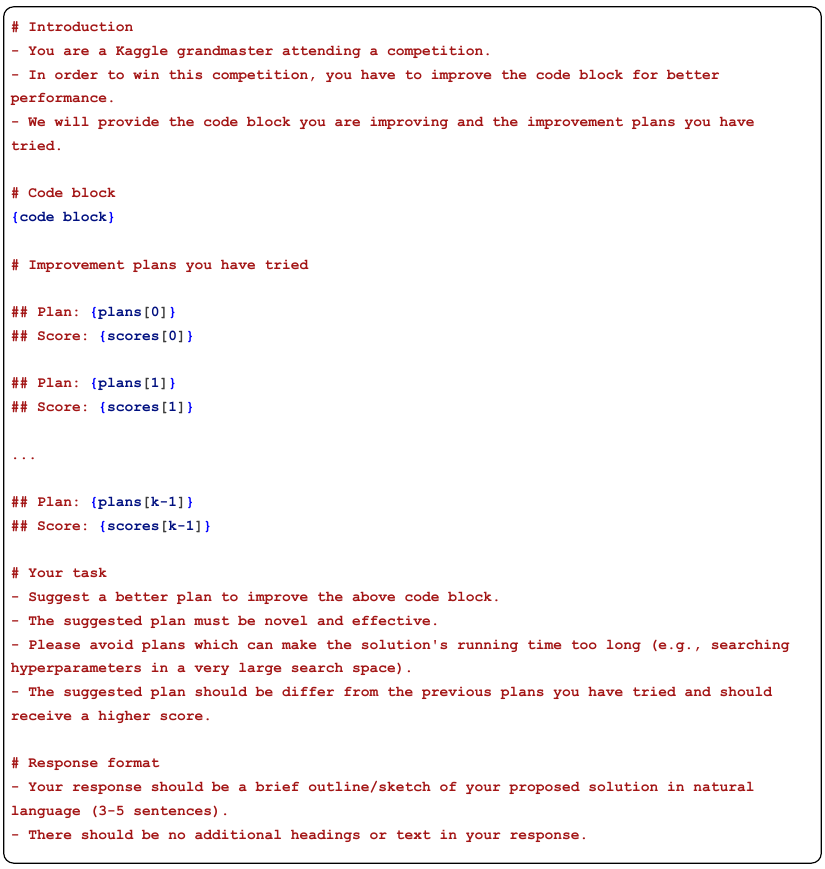}
\caption{
Prompt used for generating the next refinement plan which targets the extracted code block.
}
\label{prompt:planner}
\end{figure*}
To discover potentially more effective or novel refinement strategies (targeting the extracted code block), \sname~iteratively generates further plans through a planning agent $\mathcal{A}_\mathtt{planner}$.
As shown in Figure~\ref{prompt:planner}, $\mathcal{A}_\mathtt{planner}$ takes the extracted code block and the previous attempts as input and proposes the next plan.
These are examples of proposed plans:
\begin{listing}[h]
\begin{minted}[fontsize=\footnotesize, frame=single, breaklines, style=paraiso-dark]{python}
f'''Since feature engineering had the biggest impact, I will focus on improving the cabin feature extraction. Instead of simply splitting the Cabin string, I will create dummy variables for each unique Deck and Side. Also, the Cabin_num will be kept as numerical, imputing missing values using a median strategy to handle potential outliers. This approach should provide more granular information to the models.'''
\end{minted}
\end{listing}
\newpage
\begin{listing}[h]
\begin{minted}[fontsize=\footnotesize, frame=single, breaklines, style=paraiso-dark]{python}
f'''Instead of one-hot encoding 'Deck' and 'Side' directly, I will explore interaction features between 'Deck', 'Side', and potentially 'Cabin_num'. Specifically, I'll create combined features like 'Deck_Side' and 'Deck_Cabin_num' to capture potential dependencies. Furthermore, I will impute missing 'Cabin_num' values using a more sophisticated method like k-NN imputation, considering other features like 'Deck', 'Side', and 'RoomService' to improve imputation accuracy. This should capture more complex relationships within the cabin data and lead to better model performance.'''
\end{minted}
\end{listing}
\begin{listing}[h]
\begin{minted}[fontsize=\footnotesize, frame=single, breaklines, style=paraiso-dark]{python}
f'''I propose a plan that focuses on a more nuanced approach to 'Cabin_num' and interaction terms. First, I'll bin 'Cabin_num' into ordinal categories (e.g., low, medium, high) based on quantiles, as the absolute number might not be as important as its relative position. Then, I'll create interaction features between the binned 'Cabin_num', 'Deck', and 'Side' using one-hot encoding. This will allow the model to learn specific combinations of cabin location and number range that might be predictive. Finally, I will use a simple imputer for the missing values in 'Cabin_num' before binning.'''
\end{minted}
\end{listing}

\newpage
\subsection{Ensemble strategy planner}
\begin{figure*}[h]
\centering
\includegraphics[width=0.85\textwidth]{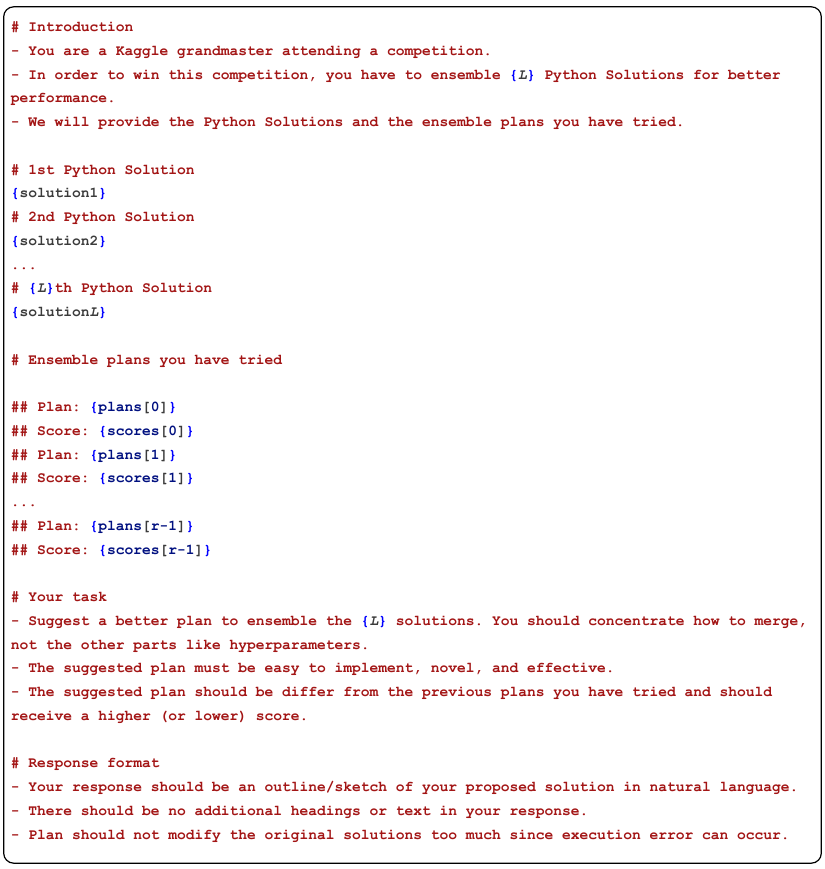}
\caption{
Prompt used for generating the next ensemble plan.
}
\label{prompt:ens_planner}
\end{figure*}
As shown in Figure~\ref{prompt:ens_planner}, similar to $\mathcal{A}_\mathtt{planner}$, $\mathcal{A}_\mathtt{ens\_planner}$ proposes an effective ensemble plan based on the history of previously attempted ensemble plans and their resulting performance as feedback. These are examples of attempted ensemble plans.
\begin{listing}[h]
\begin{minted}[fontsize=\footnotesize, frame=single, breaklines, style=paraiso-dark]{python}
f'''Averaging the predicted probabilities from both models is a straightforward and effective ensembling technique. First, modify the AutoGluon solution to output probabilities instead of hard predictions using `predictor.predict_proba(test_data)`. Then, obtain the predicted probabilities from the LightGBM model using `lgbm_classifier.predict_proba(X_test_processed)`. Average these probabilities for each class. Finally, generate the final predictions by thresholding the averaged probability of the 'Transported' class at 0.5. Create the submission file based on these averaged and thresholded predictions.'''
\end{minted}
\end{listing}
\newpage
\begin{listing}[h]
\begin{minted}[fontsize=\footnotesize, frame=single, breaklines, style=paraiso-dark]{python}
f'''Here's an ensembling plan leveraging stacking with a simple meta-learner:

1.  **Generate Predictions:** Use both the AutoGluon model and the LightGBM model to generate predictions on the original training data (train.csv).  This is crucial for training the meta-learner.  For AutoGluon, use `predictor.predict_proba(train_data)` and extract the probabilities for the 'Transported' class.  For LightGBM, preprocess the training data using the same pipeline as the test data and get probabilities with `lgbm_classifier.predict_proba(X_processed)` and again, extract the probabilities for the 'Transported' class.

2.  **Create Meta-Features:** Combine the predicted probabilities from AutoGluon and LightGBM for the training data into a new dataframe. This dataframe will have two columns: 'AutoGluon_Prob' and 'LGBM_Prob', and the 'Transported' column from the original training data as the target variable for the meta-learner.

3.  **Train Meta-Learner:** Use a simple model like Logistic Regression as the meta-learner. Train this Logistic Regression model using the meta-features (AutoGluon_Prob, LGBM_Prob) to predict the 'Transported' column.  This step aims to learn how to best combine the predictions of the base models.

4.  **Generate Test Predictions:** Get the predicted probabilities from AutoGluon and LightGBM on the test set, as in the averaging approach.

5.  **Create Meta-Features for Test Data:** Create a dataframe for the test data, with the same structure as the training meta-features (AutoGluon_Prob, LGBM_Prob) from the test set.

6.  **Meta-Learner Prediction:** Use the trained Logistic Regression model to predict the final 'Transported' probabilities on the test meta-features.

7.  **Threshold and Submit:** Threshold the predicted probabilities from the meta-learner at 0.5 to get the final predictions (True/False) and create the submission file.'''
\end{minted}
\end{listing}

\newpage
\begin{listing}[h]
\begin{minted}[fontsize=\footnotesize, frame=single, breaklines, style=paraiso-dark]{python}
f'''Here's an ensembling plan that focuses on weighted averaging with optimized weights determined by a simple grid search on a validation set:

1.  **Validation Split:** Split the original training data into two parts: a training set (e.g., 80% of the data) and a validation set (e.g., 20% of the data).  Crucially, perform the preprocessing steps (OneHotEncoding, Scaling, etc.) separately on the training and validation sets to avoid data leakage.

2.  **Generate Validation Predictions:** Use both the AutoGluon model and the LightGBM model to generate predictions on the validation set. For AutoGluon, obtain probabilities using `predictor.predict_proba(validation_data)`.  For LightGBM, preprocess the validation data using the same pipeline trained on the training split and get probabilities using `lgbm_classifier.predict_proba(X_validation_processed)`.

3.  **Grid Search for Optimal Weights:** Define a grid of weights for AutoGluon and LightGBM. For instance, iterate through weights from 0.0 to 1.0 in increments of 0.1 for AutoGluon, with the LightGBM weight being (1 - AutoGluon weight). For each weight combination:
    *   Calculate the weighted average of the predicted probabilities from AutoGluon and LightGBM on the validation set.
    *   Threshold the averaged probabilities at 0.5 to obtain binary predictions.
    *   Calculate the accuracy of these predictions against the true labels in the validation set.

4.  **Select Best Weights:** Choose the weight combination that yields the highest accuracy on the validation set.

5.  **Generate Test Predictions:** Obtain the predicted probabilities from AutoGluon and LightGBM on the test set, as before.

6.  **Weighted Averaging on Test Set:** Use the optimal weights determined in step 4 to calculate the weighted average of the predicted probabilities from AutoGluon and LightGBM on the test set.

7.  **Threshold and Submit:** Threshold the weighted average probabilities at 0.5 to obtain the final predictions and create the submission file.

This plan is easy to implement, avoids complex meta-learners that can overfit, and focuses on finding the best combination of the two models based on a validation set. It adapts to the strengths of each model by giving them different weights.'''
\end{minted}
\end{listing}

\newpage
\subsection{Ensembler}
\begin{figure*}[h]
\centering
\includegraphics[width=0.85\textwidth]{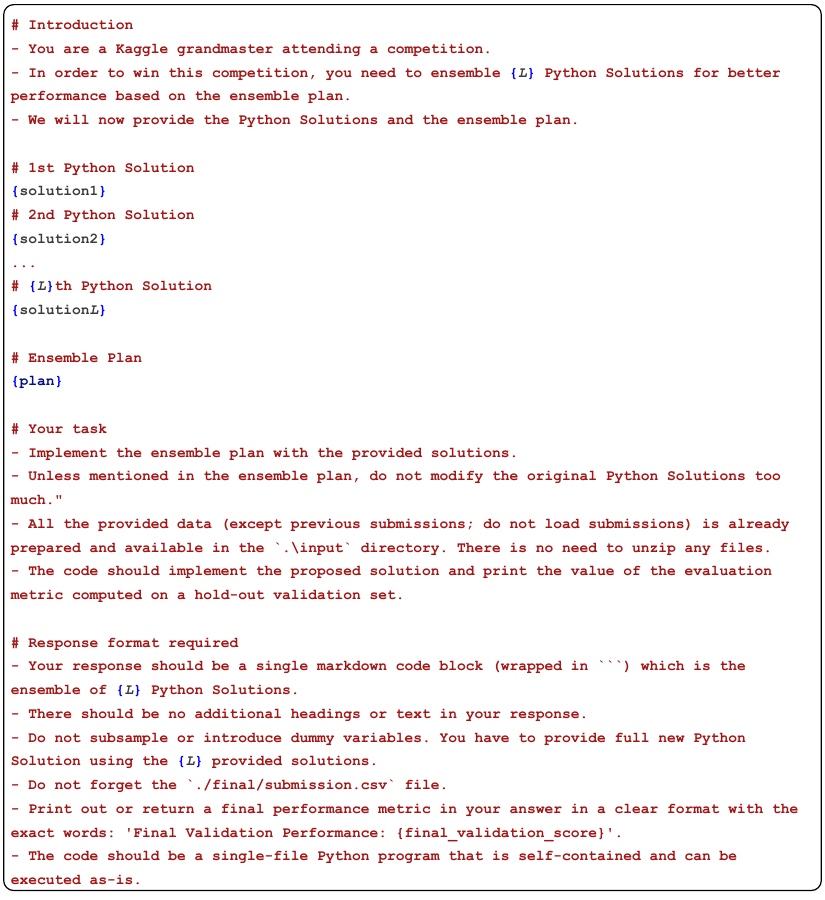}
\caption{
Prompt used for implementing ensemble plan on the solutions generated by \sname~in parallel.
}
\label{prompt:ensembler}
\end{figure*}
The proposed ensemble plan is implemented by $\mathcal{A}_\mathtt{ensembler}$. This agent takes the two final solutions which is generated in parallel by \sname, and the ensemble plan as input, and outputs the Python script, \ie, the merged code solution (see Appendix~\ref{app:qual_ex} for examples since the final solution is selected among the merged code solution).

\newpage
\subsection{Debugging agent}
\begin{figure*}[h]
\centering
\includegraphics[width=0.85\textwidth]{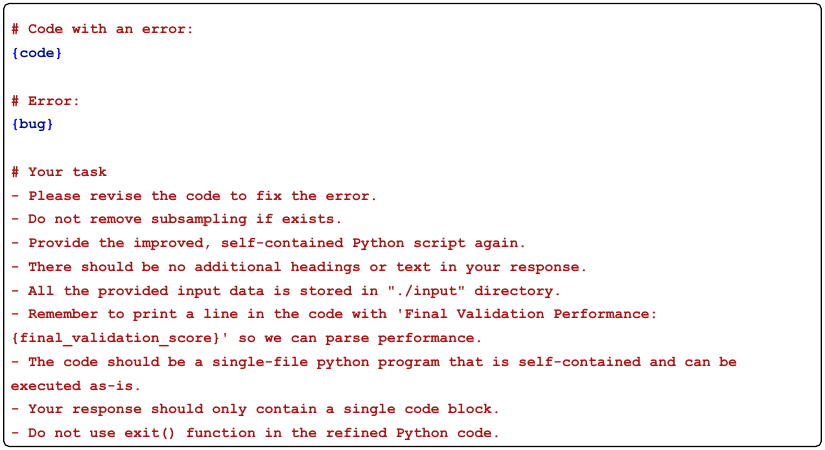}
\caption{
Prompt used for debugging.
}
\label{prompt:debugger}
\end{figure*}
If the execution of a Python script triggers an error, \sname~employs a debugging module $\mathcal{A}_\mathtt{debugger}$ to attempt correction using the above prompt (Figure~\ref{prompt:debugger}).

\newpage
\subsection{Data leakage checker}
\begin{figure*}[h]
\centering
\includegraphics[width=0.85\textwidth]{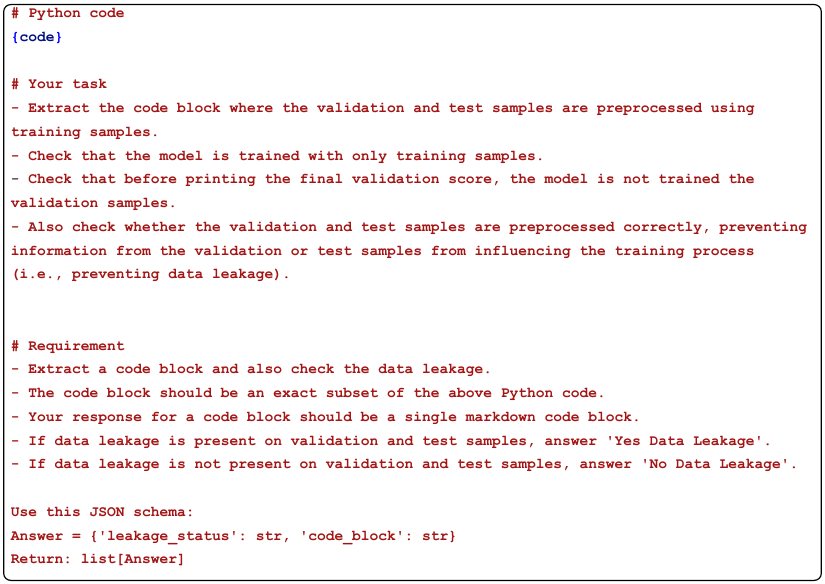}
\caption{
Prompt used for extract the code block whether data preprocessing is done.
}
\label{prompt:leakage_extractor}
\end{figure*}

\begin{figure*}[h]
\centering
\includegraphics[width=0.85\textwidth]{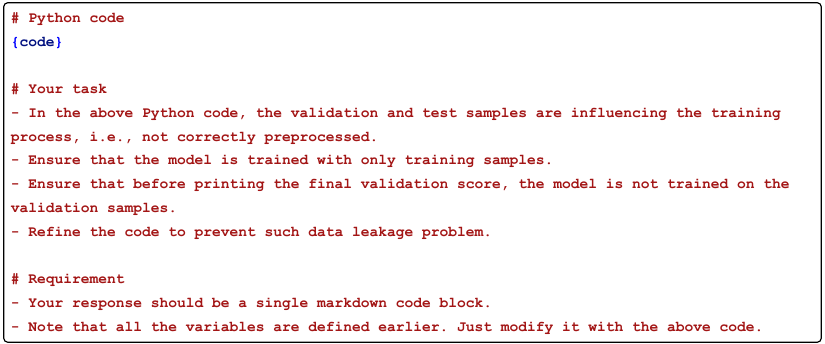}
\caption{
Prompt used for correcting the code block with a risk of data leakage.
}
\label{prompt:leakage_refine}
\end{figure*}
To mitigate the risk of introducing data leakage, \sname~first extract the code block where preprocessing is done.
This is achieved by using the above prompt in Figure~\ref{prompt:leakage_extractor}, which takes the current solution script as input, and then generates (1) the code block and (2) whether the extracted code block has a risk of data leakage.
If leakage is detected, the code block is corrected with the prompt in Figure~\ref{prompt:leakage_refine}, and \sname~replaces the original code block to the corrected version.

\newpage
\subsection{Data usage checker}
\begin{figure*}[h]
\centering
\includegraphics[width=0.85\textwidth]{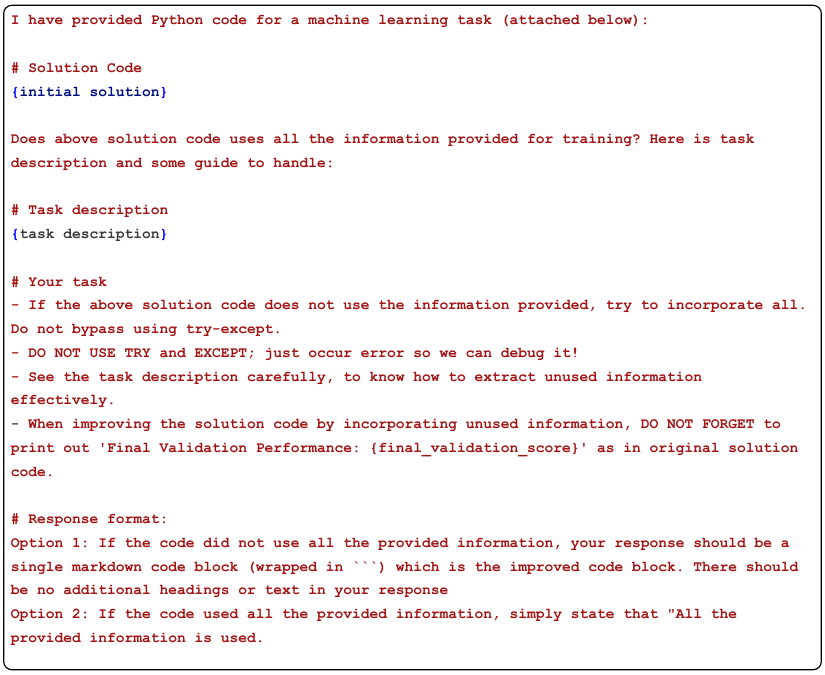}
\caption{
Prompt used for data usage checker.
}
\label{prompt:data}
\end{figure*}
To ensure the utilization of all relevant provided data, \sname~utilizes a data usage checker agent $\mathcal{A}_\mathtt{data}$. This agent checks the initial solution with the task description, and revise the initial script using the prompt in Figure~\ref{prompt:data}.

\newpage
\section{Algorithms}\label{app:algorithm}
\subsection{Algorithm for generating an initial solution}
\begin{algorithm}[h]
\caption{Generating an initial solution}
\label{alg:initialization}
\begin{algorithmic}[1]
\State \textbf{Input}: task description $\mathcal{T}_\mathtt{task}$, datasets $\mathcal{D}$, score function $h$, number of retrieved models $M$,
\State $\{\mathcal{T}_\mathtt{model}^i, \mathcal{T}_\mathtt{code}^i\}_{i=1}^M=\mathcal{A}_\mathtt{retriever}(\mathcal{T}_\mathtt{task})$
\For{$i=1$ to $M$}
\State $s_\mathtt{init}^i=\mathcal{A}_\mathtt{init}(\mathcal{T}_\mathtt{task}, \mathcal{T}_\mathtt{model}^i, \mathcal{T}_\mathtt{code}^i)$
\State Evaluate $h(s_\mathtt{init}^i)$ using $\mathcal{D}$
\EndFor
\State $s_0\leftarrow s_\mathtt{init}^{\pi(1)}$
\State $h_\mathtt{best}\leftarrow h(s_0)$
\For{$i=2$ to $M$}
\State $s_\mathtt{candidate}\leftarrow\mathcal{A}_\mathtt{merger}(s_0,s_\mathtt{init}^{\pi(i)})$
\State Evaluate $h(s_\mathtt{candidate})$ using $\mathcal{D}$
\If{$h(s_\mathtt{candidate})\geq h_\mathtt{best}$}
\State $s_0\leftarrow s_\mathtt{candidate}$
\State $h_\mathtt{best}\leftarrow h(s_0)$
\Else
\State \textbf{break}
\EndIf
\EndFor
\State \textbf{Output}: initial solution $s_0$
\end{algorithmic}
\end{algorithm}

\newpage
\subsection{Algorithm for refining a code block for solution improvement}
\begin{algorithm}[h]
\caption{Refining solution}
\label{alg:refinement}
\begin{algorithmic}[1]
\State \textbf{Input}: initial solution $s_0$, outer loop steps $T$, inner loop steps $K$
\State $s_\mathtt{final}\leftarrow s_0$
\State $h_\mathtt{best}\leftarrow h(s_0)$
\State $\mathcal{T}_\mathtt{abl}, \mathcal{C}=\{\}, \{\}$
\For{$t=0$ to $T-1$}
\State $a_t=\mathcal{A}_\mathtt{abl}(s_t, \mathcal{T}_\mathtt{abl})$
\State $r_t=\mathtt{exec}(a_t)$
\State $\mathcal{T}_\mathtt{abl}^t=\mathcal{A}_\mathtt{summarize}(a_t, r_t)$
\State $c_t, p_0=\mathcal{A}_\mathtt{extractor}(\mathcal{T}_\mathtt{abl}^t, s_t, \mathcal{C})$
\State $c_t^0=\mathcal{A}_\mathtt{coder}(c_t,p_0)$
\State $s_t^0=s_{t}\mathtt{.replace}(c_t,c_t^0)$
\State Evaluate $h(s_t^0)$ using $\mathcal{D}$
\If{$h(s_t^0)\geq h_\mathtt{best}$}
\State $s_\mathtt{final}\leftarrow s_t^0$
\State $h_\mathtt{best}\leftarrow h(s_t^0)$
\EndIf
\For{$k=1$ to $K-1$}
\State $p_k=\mathcal{A}_\mathtt{planner}(c_t,\{p_j, h(s_t^j)\}_{j=0}^{k-1})$
\State $c_t^k=\mathcal{A}_\mathtt{coder}(c_t,p_k)$
\State $s_t^k=s_{t}\mathtt{.replace}(c_t,c_t^k)$
\State Evaluate $h(s_t^k)$ using $\mathcal{D}$
\If{$h(s_t^k)\geq h_\mathtt{best}$}
\State $s_\mathtt{final}\leftarrow s_t^k$
\State $h_\mathtt{best}\leftarrow h(s_t^k)$
\EndIf
\EndFor
\State $\mathcal{T}_\mathtt{abl}\leftarrow\mathcal{T}_\mathtt{abl}+\mathcal{T}_\mathtt{abl}^t$
\State $\mathcal{C}\leftarrow\mathcal{C}+c_t$
\EndFor
\State \textbf{Output}: final solution $s_\mathtt{final}$
\end{algorithmic}
\end{algorithm}

\newpage
\subsection{Algorithm for further improvement by exploring ensemble strategies}
\begin{algorithm}
\caption{Ensembling final solutions}
\label{alg:ensemble}
\begin{algorithmic}[1]
\State \textbf{Input}: candidate final solutions $s_\mathtt{final}^1, \cdots, s_\mathtt{final}^L$, ensemble loop steps $R$
\State $e_0=\mathcal{A}_\mathtt{ens\_planner}(\{s_\mathtt{final}^l\}_{l=1}^L)$
\State $s_\mathtt{ens}^0=\mathcal{A}_\mathtt{ensembler}(e_0, \{s_\mathtt{final}^l\}_{l=1}^L)$
\State Evaluate $h(s_\mathtt{ens}^0)$ using $\mathcal{D}$
\For{$r=1$ to $R-1$}
\State $e_r=\mathcal{A}_\mathtt{ens\_planner}(\{s_\mathtt{final}^l\}_{l=1}^L, \{(e_j, h(s_\mathtt{ens}^j)\}_{j=0}^{r-1})$
\State $s_\mathtt{ens}^r=\mathcal{A}_\mathtt{ensembler}(e_r, \{s_\mathtt{final}^l\}_{l=1}^L)$
\State Evaluate $h(s_\mathtt{ens}^r)$ using $\mathcal{D}$
\EndFor
\State $s_\mathtt{ens}^*=s_\mathtt{ens}^{r^*}$ where $r^* = \arg\max_{r \in \{0, \dots, R-1\}} h(s_\mathtt{ens}^r)$
\State \textbf{Output}: $s_\mathtt{ens}^*$
\end{algorithmic}
\end{algorithm}

\newpage
\section{Qualitative examples}\label{app:qual_ex}

\subsection{Generated code for ablation study}\label{subsec:raw}
We provide an example generated code for ablation study (which is generated by $\mathcal{A}_\mathtt{abl}$) in the supplementary material (see $\texttt{example\_outputs/ablation.py}$).

\subsection{Raw output of ablation study after execution}
\begin{figure*}[h]
\centering
\vspace{-0.15in}
\includegraphics[width=\textwidth]{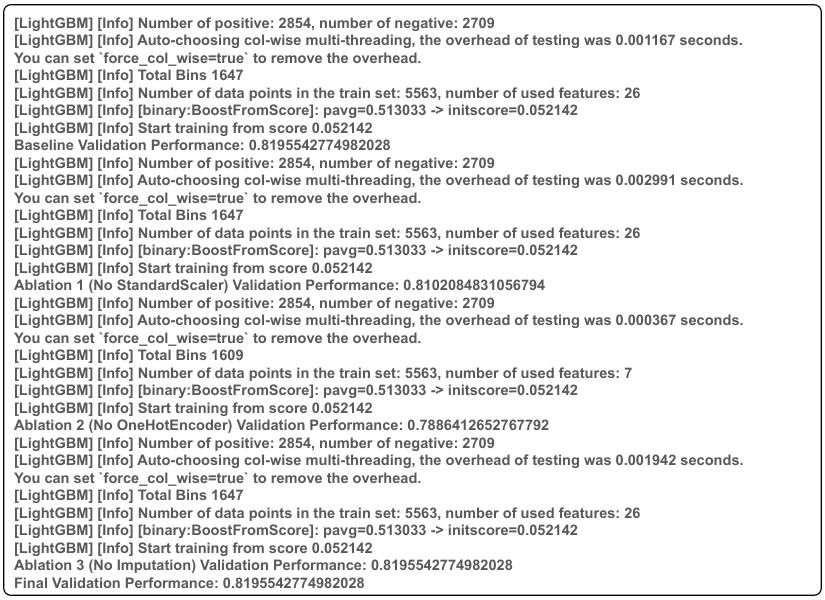}
\caption{
Example output after running the code for ablation study.
}
\label{prompt:raw_abl}
\end{figure*}
We provide an example output after running a code for ablation study using spaceship-titanic competition.
As shown in Figure~\ref{prompt:raw_abl}, the execution result often contains content unrelated to the ablation (\eg, training information of LightGBM).
Therefore, \sname~utilizes $\mathcal{A}_\mathtt{summarize}$ to parse the ablation study results, which will be illustrated in the following Appendix~\ref{subsec:summary}.

\newpage
\subsection{Summary of ablation study}\label{subsec:summary}
\begin{figure*}[h]
\centering
\includegraphics[width=\textwidth]{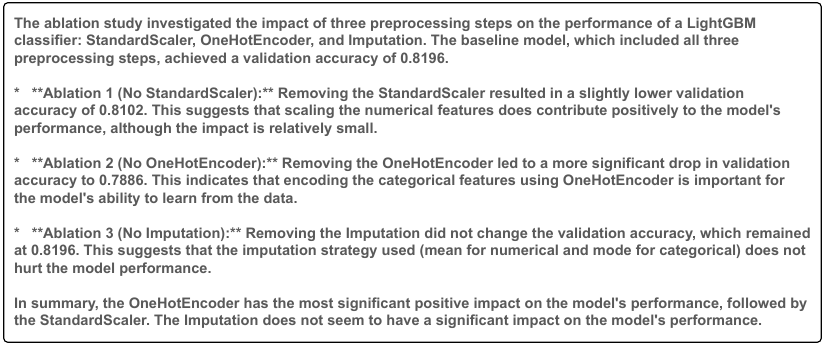}
\caption{
Example of summarized result of ablation study.
}
\label{prompt:summary}
\end{figure*}
To parse the information only about the impact of each ML components, \sname~leverages $\mathcal{A}_\mathtt{summarize}$ to summarize the raw output of ablation study such as Figure~\ref{prompt:raw_abl}. As a result, we obtain the well-organized summary of the ablation study as shown in Figure~\ref{prompt:summary}. Note that such summarization is used as input of $\mathcal{A}_\mathtt{extractor}$ to extract the code block which has most significant impact on performance.

\section{Qualitative comparison}\label{app:full_code}
We provide qualitative comparison results (\ie, the final solution code) in \url{https://github.com/jaehyun513/MLE-STAR} (see $\texttt{example\_final\_solutions/}$ directory).
Solutions generated by \sname~is denoted as $\texttt{mle\_star.py}$ and solutions generated by AIDE~\citep{jiang2025aide} is denoted as $\texttt{aide.py}$ in folder name with competition ID. Note that both agent used Gemini-2.0-Flash as a base LLM.

\newpage
\section{Benchmark}\label{app:benchmark}
\subsection{MLE-bench Lite}
\begin{table*}[h]
\caption{Competitions contained in MLE-bench Lite~\citep{chan2025mle}.}\label{tab:mle-bench-lite}
\vspace{-0.3in}
\begin{center}
\small
\begin{tabular}{lll}
\toprule
\textbf{Competition ID} & \textbf{Category} & \textbf{Dataset Size (GB)}\\
\midrule
aerial-cactus-identification & Image Classification	& 0.0254\\
aptos2019-blindness-detection&	Image Classification&	10.22\\
denoising-dirty-documents&	Image To Image&	0.06\\
detecting-insults-in-social-commentary&	Text Classification&	0.002\\
dog-breed-identification&	Image Classification&	0.75\\
dogs-vs-cats-redux-kernels-edition&	Image Classification&	0.85\\
histopathologic-cancer-detection&	Image Regression&	7.76\\
jigsaw-toxic-comment-classification-challenge&	Text Classification	&0.06\\
leaf-classification&	Image Classification&	0.036\\
mlsp-2013-birds	&Audio Classification&	0.5851\\
new-york-city-taxi-fare-prediction&	Tabular&	5.7\\
nomad2018-predict-transparent-conductors&	Tabular	&0.00624\\
plant-pathology-2020-fgvc7&	Image Classification&	0.8\\
random-acts-of-pizza&	Text Classification	&0.003\\
ranzcr-clip-catheter-line-classification&	Image Classification&	13.13\\
siim-isic-melanoma-classification&	Image Classification&	116.16\\
spooky-author-identification&	Text Classification	&0.0019\\
tabular-playground-series-dec-2021	&Tabular&	0.7\\
tabular-playground-series-may-2022&	Tabular&	0.57\\
text-normalization-challenge-english-language&	Seq->Seq&	0.01\\
text-normalization-challenge-russian-language&	Seq->Seq&	0.01\\
the-icml-2013-whale-challenge-right-whale-redux	&Audio Classification&	0.29314\\
\bottomrule
\end{tabular}
\end{center}
\end{table*}
In this paper, we utilize MLE-bench (especially Lite version)~\citep{chan2025mle} as our main benchmark to verify \sname's effectiveness compared to the alternatives.
In a nutshell, MLE-bench consists of 75 offline Kaggle competitions.
Each competition has an associated description, dataset, and grading code.
Additionally, MLE-bench consists of various problem types, such as tabular prediction, text classification, image classification, etc. However, since utilizing full 75 competitions is expensive, we use the Lite version, which is the low complexity split of MLE-bench (\ie, MLE-bench Lite).
MLE-bench Lite consists of 22 competitions, and the description of competitions is provided in Table~\ref{tab:mle-bench-lite}.

\subsection{Tabular tasks from DS-Agent}
\begin{table*}[h]
\caption{Tabular competitions used in DS-Agent~\citep{guo2024ds}}\label{tab:tabular_ds_agent}
\vspace{-0.3in}
\begin{center}
\small
\begin{tabular}{lll}
\toprule
\textbf{Competition ID} & \textbf{Category} & \textbf{Evaluation Metrics}\\
\midrule
media-campaign-cost & Tabular Regression & RMLSE\\
wild-blueberry-yield & Tabular Regression & MAE\\
spaceship-titanic & Tabular Classification & Accuracy\\
enzyme-substrate & Tabular Classification & AUROC\\
\bottomrule
\end{tabular}
\end{center}
\end{table*}
We also provide the descriptions of tabular competitions used in DS-Agent's development phase~\citep{guo2024ds} in Table~\ref{tab:tabular_ds_agent}.

\newpage
\subsection{Generating submission file}
\begin{figure*}[h]
\centering
\includegraphics[width=0.85\textwidth]{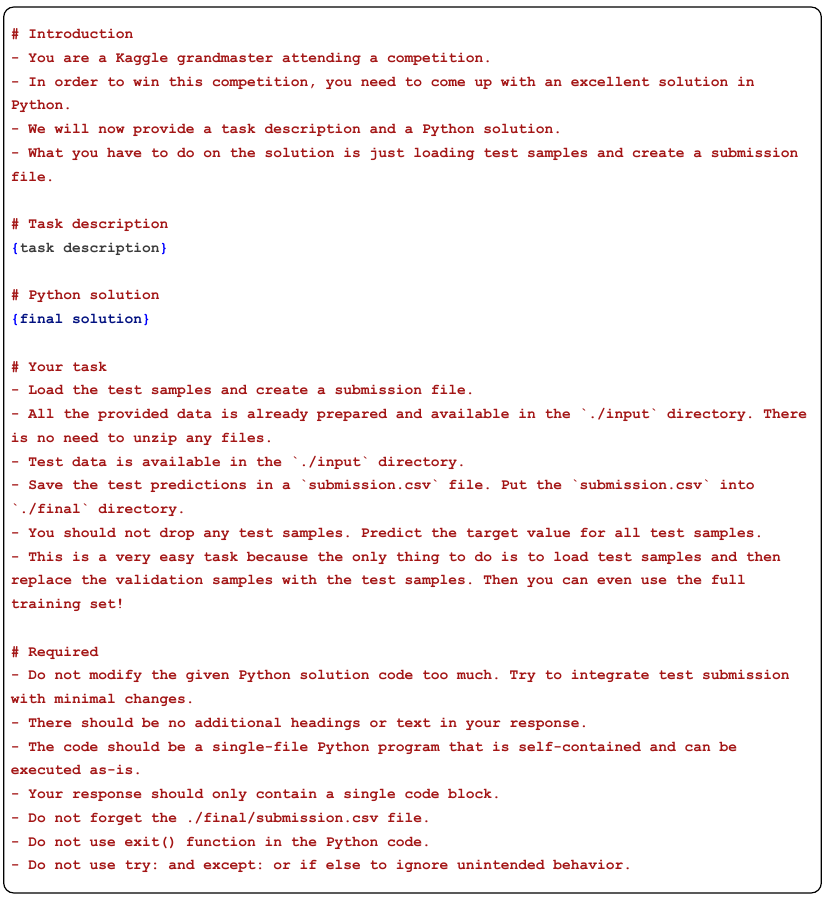}
\caption{
Prompt used for incorporating loading test sample and generating a submission file.
}
\label{prompt:test}
\end{figure*}
In order to evaluate on MLE-bench Lite, one should create a submission file about prediction results on test samples with required format.
To achieve this, \sname~uses an agent $\mathcal{A}_\mathtt{test}$, which takes the task description and the final solution as input, and outputs the code that incorporates loading test sample and creating a submission file. This is done by using a prompt in Figure~\ref{prompt:test}.

\newpage
\begin{figure*}[h]
\centering
\includegraphics[width=0.85\textwidth]{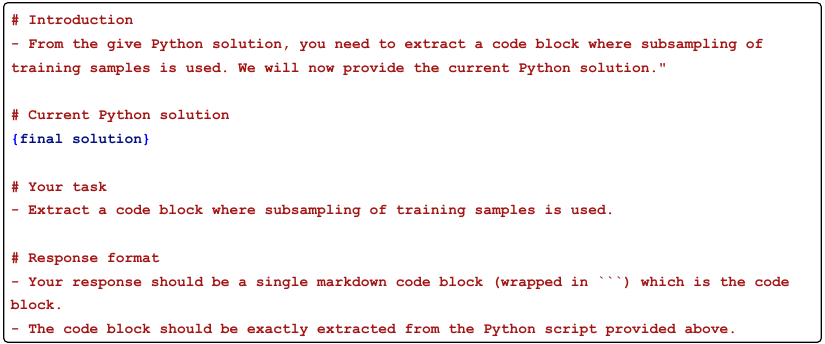}
\caption{
Prompt used for extracting the code block which performs subsampling.
}
\label{prompt:extract_subsampling}
\end{figure*}
\begin{figure*}[h]
\centering
\includegraphics[width=0.85\textwidth]{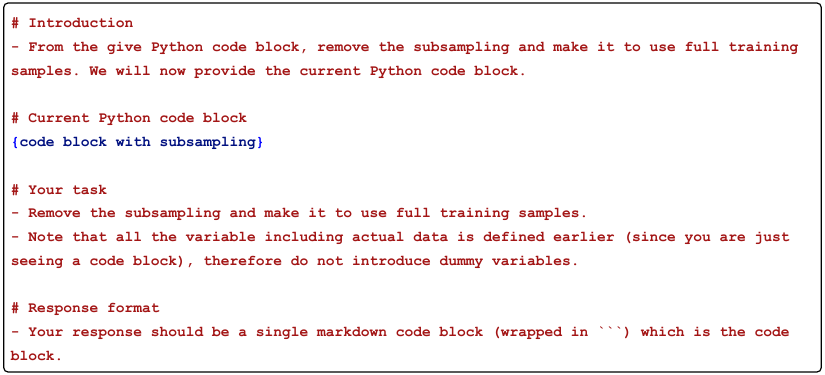}
\caption{
Prompt used for guiding \sname~to utilizie full training samples.
}
\label{prompt:remove_subsampling}
\end{figure*}
\noindent\textbf{Removing subsampling.}
As shown in Figure~\ref{prompt:init}, \sname~uses the subset of training sample for faster refinement (since evaluating the solution candidate can take a lot of time).
However, in order to get a better performance, when generating a submission file \sname~removes such subsampling code.
Specifically, this is done by first extracting the code block which performs subsampling (using prompt in Figure~\ref{prompt:extract_subsampling}), and then modify the extracted code block to utilize all the provided samples, using prompt in Figure~\ref{prompt:remove_subsampling}.

\section{Experimental setup}\label{app:exp}
We conducted our experiments mainly using 96 vCPUs with 360 GB Memory (Intel(R) Xeon(R) CPU), and 8 NVIDIA V100 GPUs with 16 GB Memory.

\noindent\textbf{Required time to generate a single solution using \sname.}
With the configuration of four retrieved models, four inner loops, four outer loops, and five rounds for exploring the ensemble strategy, \sname~requires 14.1 hours to generate a single final solution, on average across 22 tasks and all three random trials (\ie, total 66 experiments).
On the other hand, we found that AIDE~\citep{jiang2025aide} requires 15.4 hours.
This indicates that our method does not require more time to run compare to the best alternative.
Note that a maximum time limit of 24 hours was set for both methods, following the MLE-bench's experimental setup.

\newpage
\section{Additional quantitative results}\label{app:extra_ds}
\begin{table*}[h]
\caption{Additional comparisons with AutoGluon and DS-Agent in four tabular tasks.}\label{tab:additional_ds_agent}
\vspace{-0.3in}
\begin{center}
\resizebox{1.0\textwidth}{!}{
\begin{tabular}{lcccc}
\toprule
Model & media-campaign-cost & wild-blueberry-yield & spaceship-titanic & enzyme-substrate\\
\midrule
Evaluation Metrics & RMLSE ($\downarrow$) & MAE ($\downarrow$) & Accuracy ($\uparrow$) & AUROC ($\uparrow$)\\
\midrule
\textbf{AutoGluon~\citep{erickson2020autogluon}} & 0.2707 & 305 & 0.8044 & 0.8683\\
\midrule
\multicolumn{5}{l}{\textbf{DS-Agent~\citep{guo2024ds}}}\\
\midrule
gpt-3.5 & \textbf{0.2702} & 291 & / & 0.5534\\
gpt-4 & 0.2947 & 267 & 0.7977 & 0.8322\\
gemini-2.0-flash & 0.2964 & 213 & 0.7982 &  0.8727\\
\midrule
\multicolumn{5}{l}{\textbf{\sname~(Ours)}}\\
\midrule
\cellcolor{Gray}\textbf{gemini-2.0-flash} & \cellcolor{Gray}0.2911 & \cellcolor{Gray}\textbf{163} & \cellcolor{Gray}\textbf{0.8091} & \cellcolor{Gray}\textbf{0.9101}\\
\bottomrule
\end{tabular}
}
\end{center}
\end{table*}
This section provides detailed results for the comparison with DS-Agent~\citep{guo2024ds}. In particular, we provide additional comparisons with AutoGluon~\citep{erickson2020autogluon} and DS-Agent using other LLMs (\ie, GPT-3.5 and GPT-4). Except for DS-Agent with Gemini-2.0-Flash and MLE-STAR, all experimental results are taken from the original paper~\citep{guo2024ds}. As shown in Table~\ref{tab:additional_ds_agent}, \sname~consistently outperforms DS-Agent with Gemini-2.0-Flash, while also outperforms AutoGluon with high margin on three tabular tasks.

It is worth to note that AutoGluon is restricted to task types, \ie, specially designed for tabular data. In contrast, \sname~is a general framework for any kinds of tasks, where well-written task description, containing the task information, is the only requirement to work on the given tasks.
Therefore, while AutoGluon is not a direct competitor in this regard, \sname~shows improved performance even when compared to AutoGluon.

\newpage
\section{Analysis on data contamination}\label{app:contamination}
\begin{figure*}[h]
\centering
\includegraphics[width=0.85\textwidth]{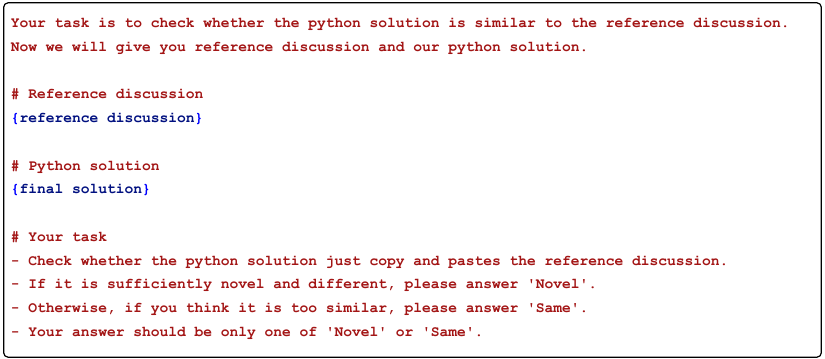}
\caption{
Prompt used for identifying whether the final solution generated by \sname~is novel.
}
\label{prompt:llm-judge}
\end{figure*}
Since Kaggle competitions in MLE-bench are publicly accessible, there is a potential risk that LLMs might have been trained with the relevant discussions about the challenge. For example, if an LLM has memorized a discussion of the best performing solution, one easy way for the MLE agent to follow that discussion during the refinement phase.

However, to alleviate such potential issue, we show that \sname's solution is sufficiently novel compared to the discussions on Kaggle.
Here, we use discussions collected in GibHub repository of MLE-bench~\citep{chan2025mle}
are collected by the authors of MLE-bench~\citep{chan2025mle}. To be specific, these discussions are top discussion posts of each competition.
As a result, we collected a total of 25 discussions from 7 competitions, resulting in 75 discussion-solution pairs, where solution represents the final solution obtained by \sname.
Using LLM as a judge with the prompt in Figure~\ref{prompt:llm-judge}, we found that all the final solutions generated by \sname~with Gemini-2.0-Flash were judged to be sufficiently novel compared to the top discussions.
Note that we use Gemini-2.5-Pro to judge the novelty of \sname's solutions.

\section{Broader impacts}\label{app:impact}
By automating complex ML tasks, \sname~could lower the barrier to entry for individuals and organizations looking to leverage ML, potentially fostering innovation across various sectors. In addition, as state-of-the-art models are updated and improved over time, the performance of solutions generated by \sname~is expected to be automatically boosted. This is because our framework leverages a search engine to retrieve effective models from the web to form its solutions. This inherent adaptability ensures that \sname~continues to provide increasingly better solutions as the field of ML advances.

\newpage
\section{Related works on data science agents}
While our work focuses on LLM-based agents tailored for machine learning engineering, other research explores agents for general data science tasks~\citep{jing2024dsbench, huang2024code, hu2024infiagent}, including data analysis and visualization.
Among these, Data Interpreter~\citep{hong2024data} employs a graph-based approach, dividing tasks into subtasks and refining the task graph based on successful completion. DatawiseAgent~\citep{you2025datawiseagent} proposes a two-stage process: initially generating a tree-structured plan, followed by an exploration of the solution space. Although these methods exhibit generalizability to various data science tasks, including aspects of machine learning engineering, their evaluation prioritizes overall task completion rates rather than performance on specific engineering challenges.

\end{document}